\documentclass[acmtog]{acmart}
\acmArticle{143}
\usepackage{booktabs} % For formal tables
\usepackage{multirow}
\usepackage{multicol}
\usepackage{color,soul}

\usepackage[ruled]{algorithm2e} % For algorithms

\SetAlFnt{\small}
\SetAlCapFnt{\small}
\SetAlCapNameFnt{\small}
\SetAlCapHSkip{0pt}

\usepackage{graphicx} 
\usepackage{subfigure}
\begin{document}
\setcopyright{acmlicensed}
\acmJournal{TOG}
\acmYear{2024} \acmVolume{43} \acmNumber{4} \acmArticle{143}\acmMonth{7}\acmDOI{10.1145/3658140}
% Title portion
\title{Interactive Character Control with Auto-Regressive Motion Diffusion Models}

% DO NOT ENTER AUTHOR INFORMATION FOR ANONYMOUS TECHNICAL PAPER SUBMISSIONS TO SIGGRAPH 2019!
\author{Yi Shi}
\affiliation{
  \institution{Simon Fraser University}
  \city{Vancouver}
  \country{Canada}} 
\affiliation{
  \institution{Shanghai AI Lab}
  \city{Shanghai}
  \country{China}} 
\email{ysa273@sfu.ca}

\author{Jingbo Wang}
\affiliation{
  \institution{Shanghai AI Lab}
  \city{Shanghai}
  \country{China}
}
\email{wangjingbo1219@gmail.com}

\author{Xuekun Jiang}
\affiliation{
 \institution{Shanghai AI Lab}
 \city{Shanghai}
 \country{China}}
\email{jiangxuekun@pjlab.org.cn}

\author{Bingkun Lin}
\affiliation{
 \institution{Xmov}
 \city{Shanghai}
 \country{China}}
\email{linbingkun014@gmail.com}

\author{Bo Dai}
\affiliation{%
 \institution{Shanghai AI Lab}
 \city{Shanghai}
 \country{China}}
\email{daibo@pjlab.org.cn}

\author{Xue Bin Peng}
\affiliation{%
 \institution{Simon Fraser University}
 \city{Vancouver}
 \country{Canada}}
\affiliation{%
 \institution{NVIDIA}
 \city{Vancouver}
 \country{Canada}}
\email{xbpeng@sfu.ca}

\renewcommand\shortauthors{Yi Shi, Jingbo Wang, Xuekun Jiang, Bingkun Lin,  Bo Dai, and Xue Bin Peng}

\begin{abstract}
     Real-time character control is an essential component for interactive experiences, with a broad range of applications, including physics simulations, video games, and virtual reality.
     The success of diffusion models for image synthesis has led to the use of these models for motion synthesis. However, the majority of these motion diffusion models are primarily designed for offline applications, where space-time models are used to synthesize an entire sequence of frames simultaneously with a pre-specified length. To enable real-time motion synthesis with diffusion model that allows time-varying controls, we propose A-MDM (Auto-regressive Motion Diffusion Model). Our conditional diffusion model takes an initial pose as input, and auto-regressively generates successive motion frames conditioned on the previous frame. Despite its streamlined network architecture, which uses simple MLPs, our framework is capable of generating diverse, long-horizon, and high-fidelity motion sequences. Furthermore, we introduce a suite of techniques for incorporating interactive controls into A-MDM, such as task-oriented sampling, in-painting, and hierarchical reinforcement learning (See Figure \ref{fig:teaser}). These techniques enable a pre-trained A-MDM to be efficiently adapted for a variety of new downstream tasks. We conduct a comprehensive suite of experiments to demonstrate the effectiveness of A-MDM, and compare its performance against state-of-the-art auto-regressive methods.
   \end{abstract}

\ccsdesc[500]{Computing methodologies~Computer graphics}
\ccsdesc[500]{Computing methodologies~Animation}
\ccsdesc[300]{Computing methodologies~Motion processing}
%
% End generated code
%

\keywords{Motion Synthesis, Diffusion Model,
Reinforcement Learning}

\let\oldtwocolumn\twocolumn
\renewcommand\twocolumn[1][]{%
	\oldtwocolumn[{#1}{
		\begin{center}
			\includegraphics[width=\linewidth]{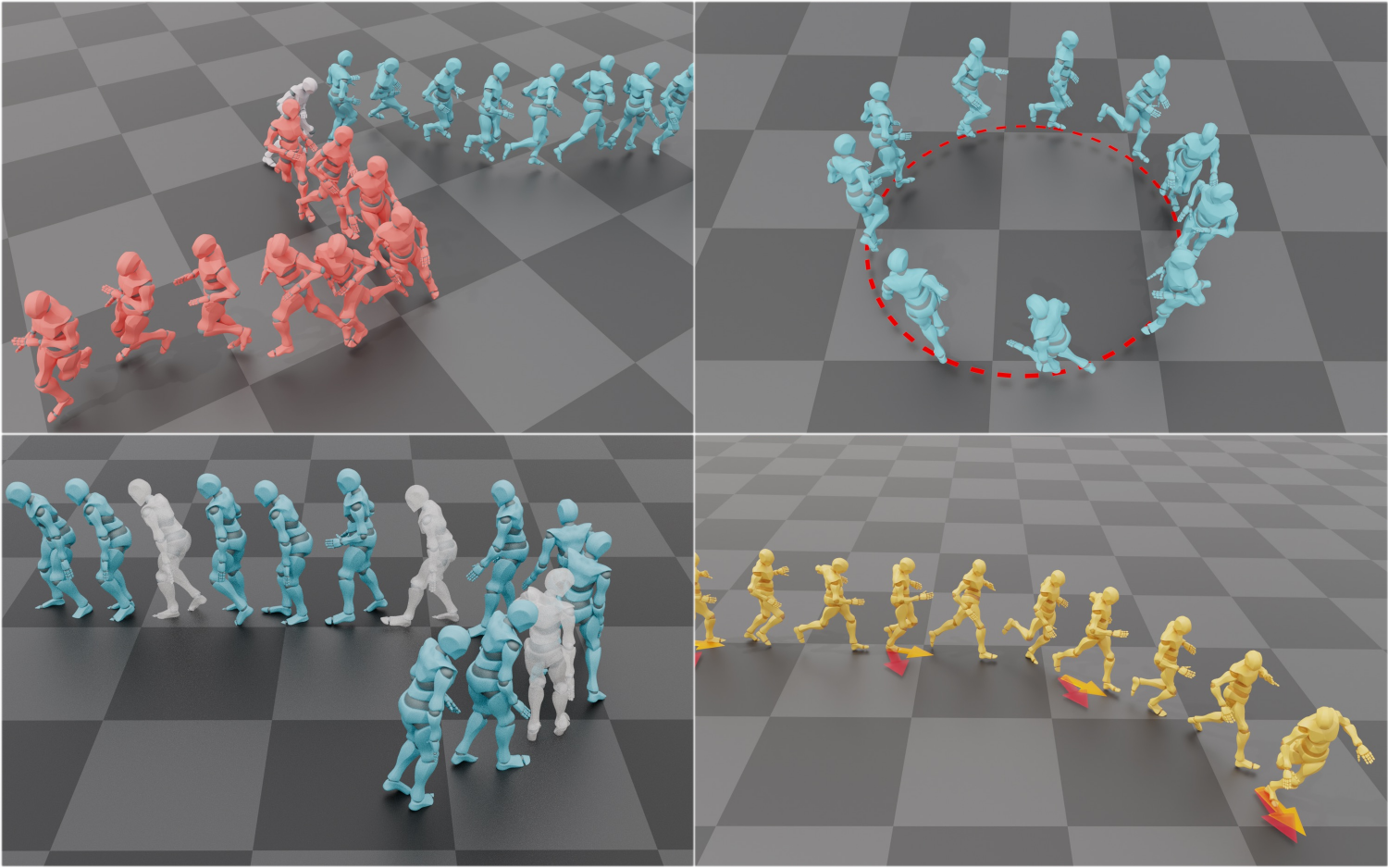}
			%\vspace {-0.4 cm}
			\captionof{figure}{\small
We present Auto-Regressive Motion Diffusion Model (A-MDM), a framework for generating high-fidelity kinematic motion sequences. Once trained, A-MDM can be reused to perform new tasks through different control strategies, such as inpainting (upper right, and lower left), and hierarchical control via reinforcement learning (lower right).}\label{fig:teaser}
\end{center}
}]
}

\maketitle

\section{Introduction}

Synthesizing high-fidelity and controllable motions for virtual characters is one of the central challenges in computer animation, with broad applications for visual effects, games, simulations, and virtual reality. The intricate nature of human motion, which encompasses a vast array of styles and tasks, presents a formidable challenge for high-quality motion generation.
In recent years, researchers have drawn on the success of data-driven methods utilized in text and image synthesis to address these challenges ~\citep{ho2020denoising,song2020denoising,rombach2021highresolution,saharia2022photorealistic,ruiz2022dreambooth}. These methods have achieved promising results in terms of motion quality and diversity. Among them, diffusion models have shown great potential in synthesizing diverse and high-quality motion sequences. However, the majority of these methods employ space-time models to generate sequences of frames simultaneously via a single diffusion process ~\citep{tevet2022human,chen2023mld,zhang2022motiondiffuse,shafir2023human,zhang2023remodiffuse,raab2023single,huang2023diffusion,dabral2022mofusion,ma2022mofusion}. This approach inherently restricts the applicability of diffusion models for tasks that require real-time interactivity.

In this work, we present Auto-regressive Motion Diffusion Models (A-MDM), a redesign of the motion diffusion model in an auto-regressive form to generate motions frame-by-frame. To improve run-time performance, our approach predicts the next frame using under 50 denoising steps, 
which is significantly less than the 1000 steps commonly used in prior methods ~\citep{tevet2022human,chen2023mld,shafir2023human}, thus enabling our model to run in real-time with modest computational resources. This design choice, combined with our incremental frame-by-frame generation approach, allows our A-MDM model to be used for motion control applications with real-time user inputs. Through both qualitative and quantitative assessments, we show that this lightweight design still retains competitive performance in terms of motion quality and diversity when compared to state-of-the-art motion synthesis models. Our model can be effectively trained using large-scale motion datasets, such as AMASS~\citep{mahmood2019amass}. Furthermore, we also explore techniques for combining A-MDM with high-level policies to create character controllers for a variety of downstream tasks, such as joystick control, target reaching, and trajectory tracking.

Once trained, A-MDM can act as a flexible base model that can be used in a variety of downstream applications. We present a suite of techniques for controlling A-MDM for downstream tasks, including task-oriented sampling, motion in-painting, keyframe in-betweening, and hierarchical control. The key contributions of this work include:
\begin{itemize}
    \item We propose A-MDM, an auto-regressive motion diffusion model, which is more amenable for real-time interactive character control. A-MDM can generate more diverse motion sequences than previous VAE-based auto-regressive models ~\citep{ling2020character,rempe2021humor}.
    \item We present a suite of methods for controlling A-MDM to synthesize motions for new tasks without any additional training or fine-tuning of the model.
    \item We present a hierarchical reinforcement learning approach for controlling A-MDM, which is able to train hierarchical policies for new downstream tasks.
    
\end{itemize}

\section{Related Works}
Developing procedural methods that can automatically synthesize life-like animations for virtual characters has been one of the fundamental problems in computer graphics. A large suite of techniques has been proposed to tackle this problem, which can be broadly categorized into kinematic and physics-based techniques ~\citep{peng2018deepmimic, peng2021amp, peng2022ase, tevet2022human, li2022ganimator, rempe2021humor}.

Since our work falls into the class of kinematic methods, our discussion will focus on the most closely related work on kinematic motion generation. The core of our system is an auto-regressive motion diffusion model trained on a dataset of motion clips, which can then reused by a variety of control techniques to synthesize motions for new downstream tasks.

\subsection{Kinematic Motion Generation}
Kinematic techniques typically follow one of two main paradigms: space-time models and auto-regressive models.Space-time models produce an entire motion sequence simultaneously ~\citep{yan2019convolutional,tevet2022human,chen2023mld,li2022ganimator}. Alternatively, auto-regressive methods generate a motion sequentially frame-by-frame ~\citep{fragkiadaki2015recurrent,martinez2017human,ling2020character,rempe2021humor}, allowing greater responsiveness to time-varying objectives, and enabling these models to be more easily used for real-time applications.
In this paper, we focus on auto-regressive generation, which can produce motions for real-time interactive applications. In this section, we will provide a review of both space-time and auto-regressive methods.

\paragraph{Auto-Regressive Models}
Machine learning techniques for auto-regressive motion generation have often used CNNs, LSTMs, and MLPs to sequentially predict the next frame conditioned on previously generated frames ~\citep{fragkiadaki2015recurrent, li2017auto, martinez2017human, pavllo2018quaternet, gopalakrishnan2019neural, aksan2019structured}. However, applying these models directly to generate long-horizon motions often leads to drift or convergence to a mean pose, resulting in unnatural motions. Researchers have introduced custom architectures to address these issues, which can improve motion quality and diversity. For instance, phase-functioned neural networks dynamically blend between different model parameters based on the motion phase to generate coherent long-horizon motions ~\citep{holden2017phase}. However, this dependency on phase undermines its ability to model acyclic behaviors, and behaviors that do not progress according to a well defined phase variable. 

Variational auto-encoder models (VAE) have been a popular class of generative model for motion synthesis. These methods leverage a VAE framework to learn a motion manifold from motion data, which can enhance diversity and generalization in the generated motions ~\citep{kingma2013auto,ling2020character,rempe2021humor}. Motion VAE (MVAE) employs a Mixture-of-Expert (MoE) network alongside a VAE model to generate diverse motions ~\citep{ling2020character}.

Humor models the manifold of transitions between two adjacent frames via a VAE model ~\citep{rempe2021humor}. ~\citet{hassan2021stochastic} utilizes a VAE model to synthesize long-horizon motions of human-scene interactions. In our work,  we introduces an auto-regressive diffusion model, which leverages the expressive power of diffusion model to better represent the complex multi-modal nature of human motions. This expressiveness enables our model to produce more diverse and higher fidelity motions compared to previous auto-regressive models.

\paragraph{Space-Time Models}
Unlike auto-regressive models, space-time models generate a sequence of motion frames simultaneously. As result, these models are typically more suitable for offline applications. 
Space-time models based on CNNs ~\citep{yan2019convolutional, wang2021scene}, and transformer are popular choices in prior studies ~\citep{petrovich2021action, wang2022towards,kim2022conditional}, as these architectures are able to capture temporal correlation across long motion sequences. VAE ~\citep{kingma2013auto} and GAN ~\citep{goodfellow2020generative} frameworks have been used to trained space-time models. 

Recent works have applied diffusion model to space-time motion synthesis models ~\citep{tevet2022human,zhang2022motiondiffuse}.
However, a common limitation of diffusion models is the significant computational cost and time needed for sampling, which can preclude effective use in real-time applications. \citet{chen2023mld} attempted to address this challenge by building a diffusion model in a compact VAE-encoded latent space. While their method demonstrated some real-time generation capabilities, the time required to generate a single frame is still too slow for real-time applications. In this work, we propose a lightweight auto-regressive diffusion model with specialized design decisions to speedup generation of motions in an auto-regressive fashion, which then enables our A-MDM model to synthesize high-fidelity motions in real-time and respond to interactive control inputs.

\subsection{Latent-Space Models}
%Latent-space model typically starts by learning a latent motion representation, which can then be used to generate motions of new tasks. 
Latent-space models typically leverage a learned latent motion representation, which can then be used to generate new motions for downstream tasks.  Commonly used approaches for learning such latent-space models utilize generative modeling techniques, such as VAEs and GANs. MVAE generates motions using latent motion representations learned through a conditional variational auto-encoder ~\citep{ling2020character}. \citet{won2022physics} and \citet{yao2022cvae} leverage model-based RL methods to train physics-based motion VAEs. Once a latent space of motions has been constructed, these frameworks then train task-specific high-level controllers to sample the appropriate latent codes for solving new tasks. In practice, VAE models tend to generate lower-quality results compared to diffusion models in the image synthesis domain ~\citep{ho2020denoising, gustav2021vaeblur}, and similar issues have also been observed for motion generation with space-time model ~\citep{tevet2022human, petrovich22temos, Guo_2022_CVPR}. 
In this work, we show that with the appropriate design decisions, diffusion models can effectively generate high-quality motions for real-time responsive character control.

GAN-based models learn a latent motion representation by optimizing a variational approximation of the divergence between the dataset and samples from the model ~\citep{peng2022ase, peng2021amp, tessler2023calm, li2022ganimator, BarsoumCVPRW2018, men2022ganclass}.

This is done by training an adversarial discriminator to distinguish samples from the dataset and samples from the model, thereby encouraging the model to produce samples that resemble the data. However, GAN training tends to be unstable and susceptible to mode collapse, requiring a myriad of heuristics to stabilize training.

In this paper, our method leverages diffusion models for motion synthesis, which has been shown to be much more stable and easier to train compared to GANs. It is also known to achieve better generation quality and data distribution coverage than adversarial methods in other domains, such as image synthesis ~\citep{dhariwal2021}.

To apply VAE and GAN models to new tasks, hierarchical controllers are often trained to select the appropriate behaviors from the learned latent space that enable a character to fulfill the desired task objectives \citep{tessler2023calm,peng2022ase,ling2020character}. 

We show that A-MDM is also amenable to hierarchical control, where a task-specific high-level controller can be trained to steer the denoising process in order to generate motions that satisfies the desired task objectives.
\section{Method}

\begin{figure*}[h]
    \centering
    \includegraphics[width=1.0\textwidth]{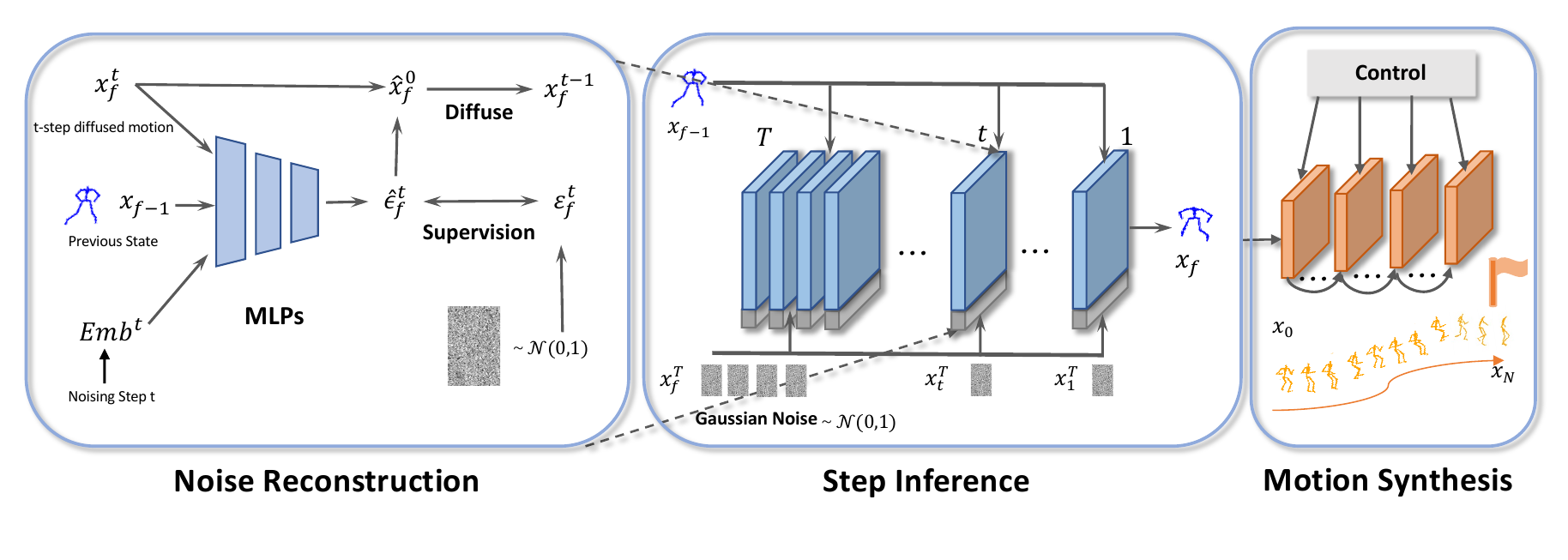}\\
    \vspace{-0.9cm}
    \caption{\textbf{Framework of our A-MDM.} Our A-MDM is trained following DDPM~\cite{ho2020denoising}. During training, the goal of our A-MDM is to reconstruct the sampled noise vector $\epsilon^t_f$ at each step. After training, our A-MDM is capable of generating long-horizon motion with arbitrary lengths under different controlling strategies in an autoregressive manner.}
    \label{f1_pipeline}
\end{figure*}

Our framework consists of two key components: a base autoregressive motion diffusion model (A-MDM) and an RL-based task controller. The base A-MDM predicts the next motion frame $f$ based on the previous frame $f-1$. This auto-regressive procedure enables A-MDM to synthesize long motions of arbitrary lengths in real time. To achieve task-based character control, we utilize a task-specific high-level controller, which is trained using reinforcement learning to direct the base A-MDM model to produce motions that fulfill a given task objective. This framework enables our system to generate high-quality, task-oriented motions for a wide range of applications. 
%To further enhance the diversity and flexibility of the generated motions, we incorporate in-painting techniques such as those utilized in \cite{tevet2022human, shafir2023human}. These techniques enable real-time motion control and result in a more realistic and natural appearance for the generated motions.
Furthermore, we demonstrate that inpainting techniques can also be incorporated into our auto-regressive model, which enables users to directly specify desired characteristics of the generated motions. We show that the model is able to generate natural and responsive motions that adhere to the target specifications provided by the user. 

\subsection{Auto-regressive Motion Diffusion Model}\label{Method_AMDM}
In this section, we will present the details of our auto-regressive motion diffusion model. Given the state of the character at frame $f-1$,  A-MDM models the distribution of possible character states at frame $f$. The training procedure of A-MDM follows DDPM, where the model is trained to model the distribution of the motion data by denoising Gaussian noise introduced by a forward diffusion process. Once trained this model can then produce diverse, high-fidelity, and long-horizon motion sequences by sampling from the reverse diffusion process \citep{ho2020denoising}. 

\paragraph{Motion Representation}\label{Method_MR}
Given the initial state $x_0$, A-MDM can generate a motion sequence $X=\{x_1, x_2,.., x_F\}$ with arbitrary length $F$ in an auto-regressive manner. The state $x_f$ at frame $f$ is represented by features consisting of: planar linear velocity of the root ($d_x\in\mathbb{R}$, $d_y\in\mathbb{R}$), angular velocity ($d_r\in\mathbb{R}$) around the up-axis, joint positions ($j_p\in{\mathbb{R}^{j\times3}}$), joint velocities ($j_v\in{\mathbb{R}^{j\times3}}$), and joint orientations in the 6D representation ($j_o\in{\mathbb{R}^{j\times6}}$)~\citep{zhou20196d}. The root features $d_x$, $d_y$, and $d_r$ are recorded in the character's local coordinate frame.
%This formulation of root features results in a higher frequency of occurrence of values, thereby improving the generalizability of an auto-regressive motion model ~\citep{ling2020character}. 

\paragraph{Pipeline Overview}\label{Method_PO}
Given an input character state, $x_{f-1}$, A-MDM predicts a potential next state $x_{f}$ in an auto-regressive manner. The model is trained using a procedure based on DDPM \cite{ho2020denoising}, which includes a forward diffusion process and a backward reconstruction process. During the forward diffusion process, Gaussian noise is introduced at each diffusion step $t$ into the next state $x_{f}^t$. This process is applied for $T$ steps until $x_{f}^T$ converges to an isotropic Gaussian distribution. The backward reconstruction process again consists of $T$ steps, representing a reversed procedure where a conditional denoising model is trained to gradually remove the noise in $x_f^t$ conditioned on the previous state $x_{f-1}$, yielding the final prediction $x_{f}$. 

During training, Gaussian noise is added progressively through a Markov process to the ground truth character state $x_{f}$.  For a single forward step $t$, the noising process is given by:
\begin{equation}\label{e1_diff_single}
q(x_{f}^{t}|x_{f}^{t-1})=\mathcal{N}(x_{f}^t; \sqrt{1-\beta^t}x_{f}^{t-1},\beta^tI),
\end{equation}
where $\beta{^t}\in(0,1)$ is a hyper-parameter that determines the noise schedule. The whole forward diffusion process can be described according to: 
\begin{equation}\label{e2_diff_whole}
q(x_{f}^{1:T}|x_{f}^{0})=\prod_{t=1}^T{q(x_{f}^{t}|x_{f}^{t-1})},
\end{equation}
We denote the noise applied to state $x_{f}$ after $t\in[1,T]$ steps as $\epsilon_f^t$. The denoising process is performed using a neural network $p_\theta$. The network takes as input the perturbed target state $x_f^{t}$, which is produced by applying noise $\epsilon_f^t$ from the standard Gaussian distribution to the state $\hat x_{f}^0$ at denoising step $t+1$. Additional input to the network includes the previous state $x_{f-1}$, the time embedding $e^t$ at time step $t$. The denonising network then predicts the noise vector $\hat\epsilon_f^t$ that was applied to the original input. An MSE loss is applied between the predicted noise and the ground truth noise $\epsilon_f^t$:
\begin{equation}\label{e3_train_loss}
L_t^{simple}(x) = \mathbb{E}_{t\sim[1:T],x_f^0,\epsilon_f^t}||\epsilon_f^t-p_\theta(x_{f-1}, x_{f}^{t}, {e}^t)||_2 .
\end{equation}
During inference, the reverse denoising process starts from Gaussian noise $\epsilon^T_f$. At each diffusion step $t$, the denoising network predicts the noise $\hat{\epsilon}^t_f$, which is used to reconstruct the original state ${\hat{x}^t_f}$. Then, ${\hat{x}^t_f}$ serves as the input to the next denoising step. The final predicted state $\hat{x}_{f}$ is used as the output ${\hat{x}^0_f}$ after $T$ denoising steps.

\begin{figure*}[t!]
    \subfigure{\includegraphics[width=3.5cm]{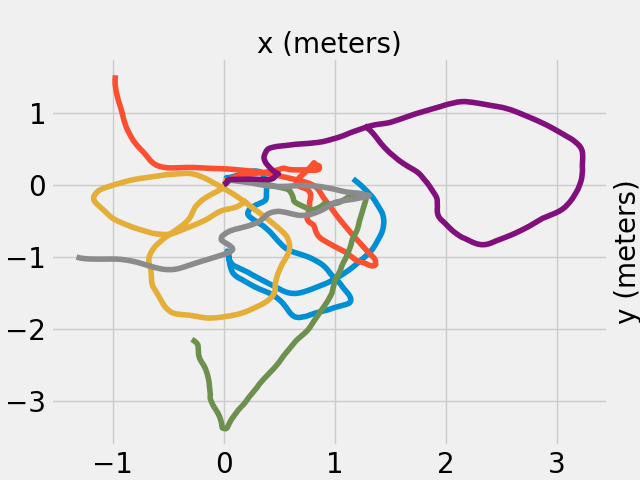}}
    \subfigure{\includegraphics[width=3.5cm]{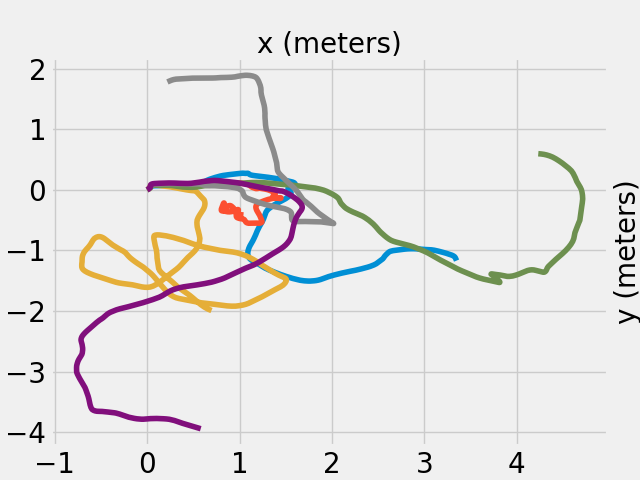}}
    \subfigure{\includegraphics[width=3.5cm]{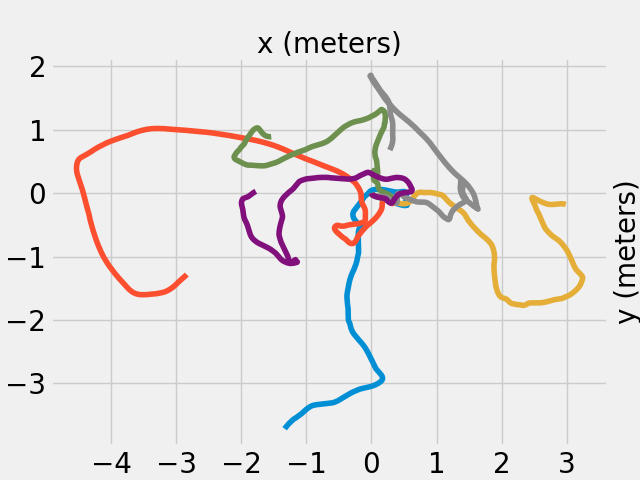}}
    \subfigure{\includegraphics[width=3.5cm]{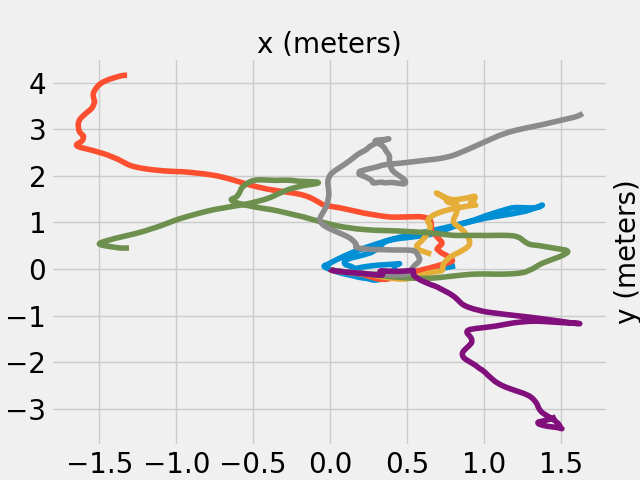}}
    \subfigure{\includegraphics[width=3.5cm]{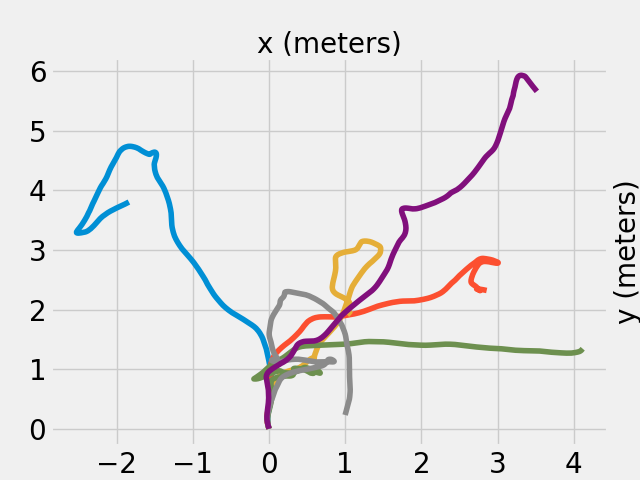}}
    \vfill
    \subfigure{\includegraphics[width=3.5cm]{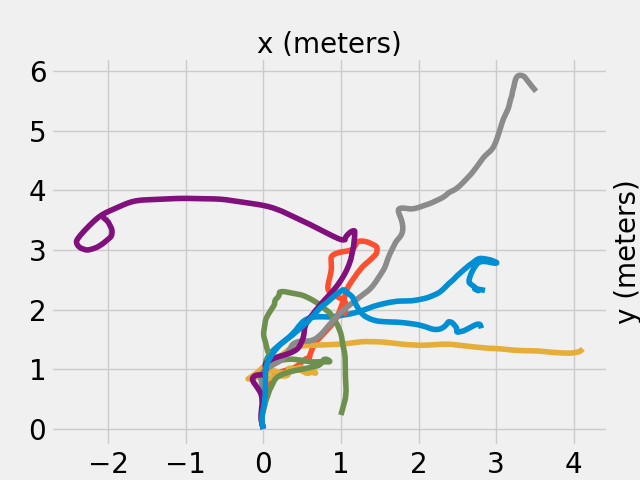}}
    \subfigure{\includegraphics[width=3.5cm]{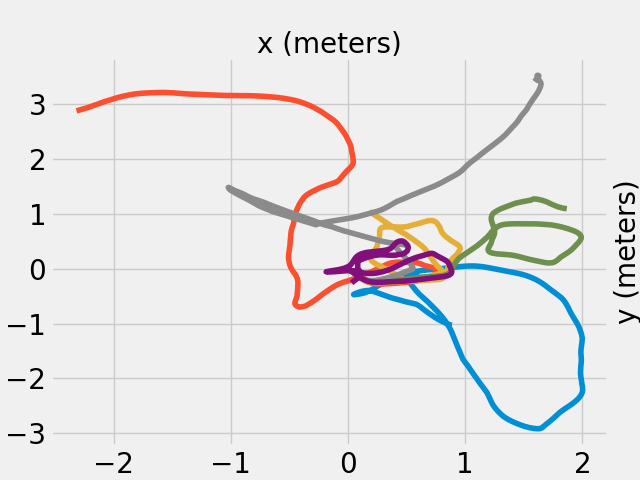}}
    \subfigure{\includegraphics[width=3.5cm]{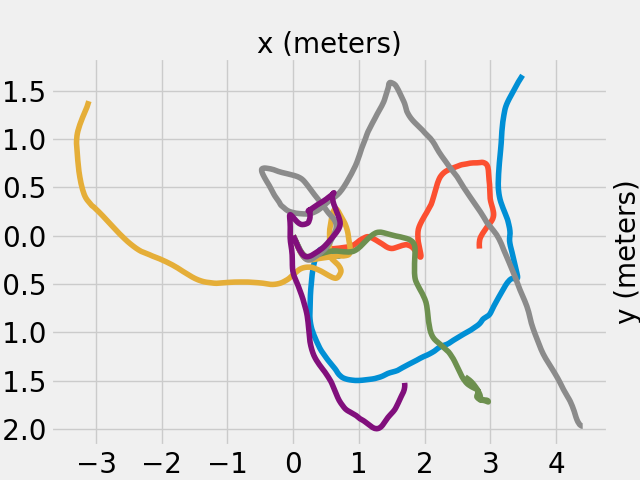}}
    \subfigure{\includegraphics[width=3.5cm]{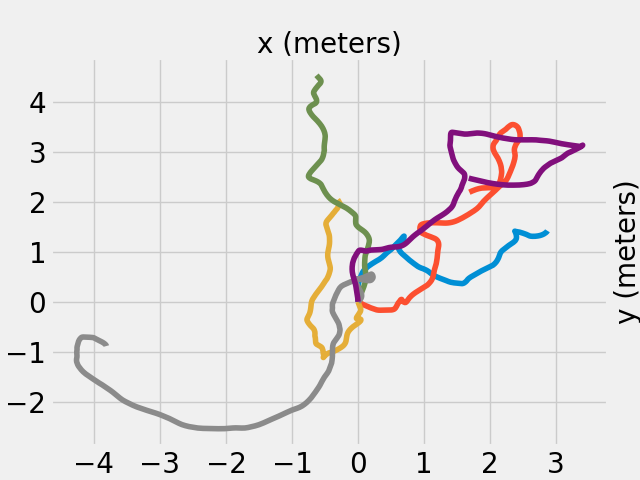}}
    \subfigure{\includegraphics[width=3.5cm]{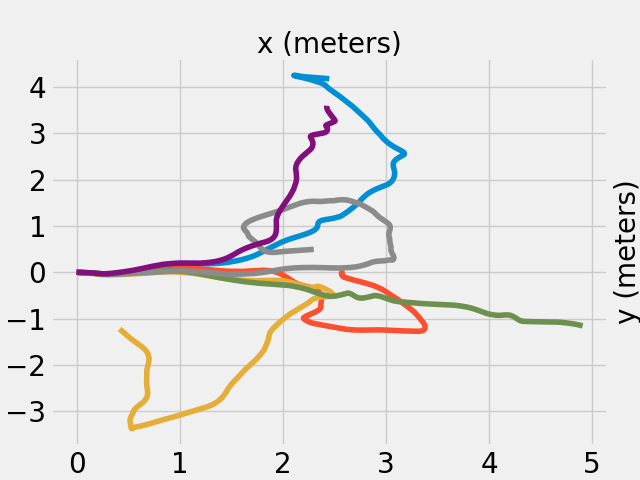}}
    \caption{\textbf{Trajectories for Target Reaching.} We show that our framework is capable of generating diverse motion trajectories, with the same initial state and target goals. }
    \label{fig:rl_target}
\end{figure*}
\subsection{Scheduled Sampling with Diffusion Model}\label{scheduled_sampling}
In Section~\ref{Method_PO}, we discussed the basic training and inference procedure for A-MDM. However, generating motion with an auto-regressive model is susceptible to drift due to error accumulation. This phenomenon can lead to catastrophic failures when generating long-horizon motions. To mitigate drift, prior methods leverage student forcing within the framework of schedule sampling \cite{ling2020character,rempe2021humor}. In comparison to teacher forcing, where both the input and the target output are sampled from the training data, student forcing uses the model's own predicted state from a past frame $\hat{x}_{f-1}$ as the input to predict the subsequent state $\hat{x}_{f+1}$. This process can be repeated over multiple steps $N_r$. Scheduled sampling enables an auto-regressive model to compensate for drift due to errors in its previous predictions. This can then enable auto-regressive models to generate long-horizon motions with arbitrary length, which is essential for online interactive applications. In this work, we observed that scheduled sampling is also crucial for mitigate drift with A-MDM. To implement student forcing, for every frame $x_{f}$ in a batch of samples, we  compute the loss described in Equation \ref{e3_train_loss}, and then use the reconstructed $\hat{x}_f^t$ as the next frame for reconstructing subsequent frames. Following the same procedure, we can sample $N_r$ subsequent frames auto-regressively from the model $\{\hat{x}_{f+1}, \hat{x}_{f+2}, ... , \hat{x}_{f+N_r}\}$. The generated frames are then used as the input conditioning for computing the training loss for subsequent frames instead of the ground-truth frames from the dataset.

\subsubsection{Network Architecture}
To enable real-time motion control, our framework leverages a lightweight network architecture for A-MDM. Unlike prior work, which often uses computationally expensive architectures, such as Transformers \citep{tevet2022human,chen2023mld}, our A-MDM model features a streamlined architecture centered around a lightweight Multi-Layer Perceptron (MLP), as illustrated in Figure~\ref{f1_pipeline}. The use of a diffusion model significantly enhances our model's capacity, enabling the generation of diverse and high-quality motions using a simple model architecture. We demonstrate that using MLPs in diffusion models with fewer denoising steps is sufficient for synthesizing high-quality motions, outperforming prior auto-regressive models, such as MVAE, in both motion fidelity and diversity. Our model is composed of 10 fully-connected layers, with each layer containing 1024 hidden units with SiLU activations followed by a layer norm. The input to this model includes the state at the previous step $x_{f-1}$, the perturbed target state $x_{f}^t$, and the time embedding $e^t$ of the diffusion step $t$.

\paragraph{Real-time Design Decisions}\label{Method_SU}
In diffusion models, a common practice is to use a large number of diffusion steps (\textit{e.g.}, 1000 steps as used in DDPM). However, for auto-regressive motion generative models, like A-MDM, this can lead to slow generation times, rendering them impractical for real-time applications. Our experiments demonstrate that achieving high-quality motion generation with A-MDM is possible using a smaller number of diffusion steps $T$. This is in part attributed to the smaller dimensional output for a single frame compared to an entire motion sequence. Our empirical findings suggest that very few steps (\textit{e.g.}., 10-50) are sufficient for motion generation tasks.

\section{Motion Synthesis}~\label{Method_MS}
Once trained, A-MDM is able to generate long-horizon motions with unbounded lengths by auto-regressively sampling from the diffusion model at each time step. We present four methods for long-term motion generation: random sampling, task-oriented sampling, conditional inpainting, and hierarchical control. 

\paragraph{Random Sampling:} The first method, random sampling involves sampling from A-MDM without imposing any constraints. We show that our model is able to generate diverse multi-modal motions starting from a fixed initial state.

\paragraph{Task-Oriented Sampling:} Task-oriented sampling involves generating future states through beam search with a user-defined target score function. Several randomly sampled candidates trajectories are generated and evaluated by the score function, and the first frame from the best trajectory is selected as the next frame. This task-oriented sampling approach enables A-MDM to perform tasks, such as goal-reaching, without requiring further training. However, it is essential to note that task-oriented sampling often introduces unpredictable behaviors, as also observed in prior work \citep{ling2020character}, where sub-optimal behaviors such as running in circles around a goal may arise. It also lacks support for the user to specify precise fine-grained constraints, beyond defining how the target score is calculated.

\paragraph{Conditional Inpainting:} The third method employs inpainting techniques to generate motions that adhere to user  specifications. This technique has been highly effective for image synthesis \citep{lugmayr2022repaint}, and space-time motion models \citep{tevet2022human}. With inpainting, users can specify trajectories and positions for various joints in the character's body, and A-MDM is then used to generate a plausible full-body motion that adheres to those constraints.

\paragraph{Hierarchical Control:} Finally, we show that A-MDM can also be incorporated into a hierarchical reinforcement learning framework to train task-specific high-level controllers on top of the base A-MDM model. This enables our framework to create controllers for accomplishing new tasks, while producing higher motion quality and robustness compared to task-oriented sampling and inpainting.

\subsection{Synthesis via Random Sampling}
Given the initial character state, our A-MDM can generate plausible motions for future frames by sampling auto-regressively from the diffusion model, enabling the generation of diverse long-horizon motion sequences in real time. The diverse trajectories of synthesized motion from different initial states are shown in Figure~\ref{fig:rl_target}. Our supplementary video also demonstrates the capability of our A-MDM to transition from a stationary pose to diverse behaviors, such as walking, running, and jumping. When presented with different initial states, A-MDM is capable of producing motions that are consistent with the initial state.  For instance, it can generate hopping motions when initialized with a hopping pose. These instances showcase the model's ability to learn the distribution of plausible motions associated with specific states. Our supplementary video offers additional examples, providing a comprehensive showcase of the diverse motions that can be generated by A-MDM.

\begin{figure}[t!]
    \centering
    \includegraphics[width=0.5\textwidth]{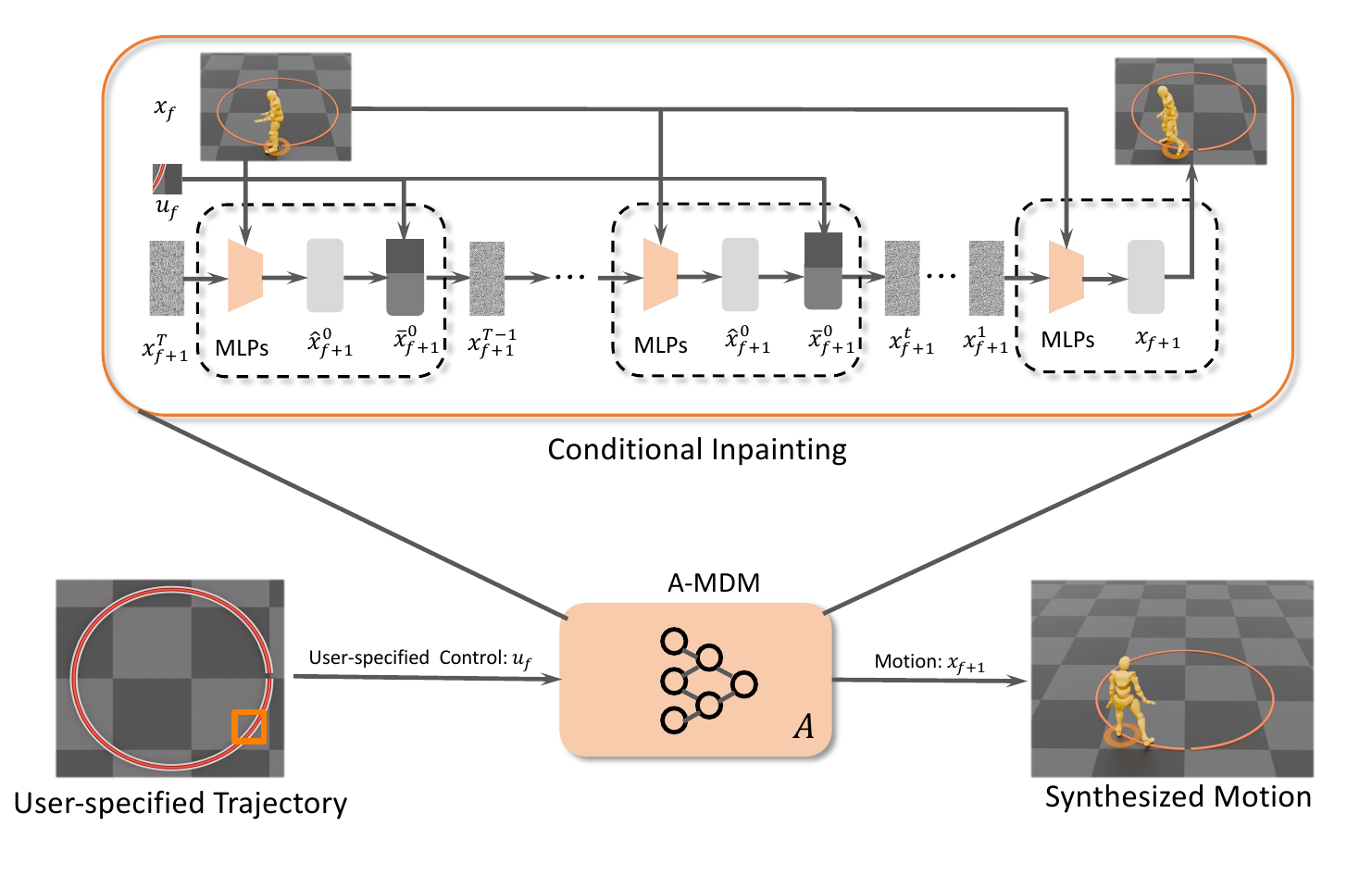}\\
    \vspace{-0.4cm}
    \caption{Inpainting can generate seamless motion transitions between user-specified motions and arbitrary character states. We introduce a series of buffer frames where inpainting stops at an early diffusion step. While playing out the user-imposed target motion, inpainting is done until the last denoising step.}
    \label{fig:inpt}
\end{figure}

\subsection{Synthesis via Task Oriented Sampling}
Random sampling alone cannot generate motions that conform to user commands. Therefore, following the sampling-based control method presented by ~\citet{ling2020character}, we implement a task-oriented sampling technique to generate motions that follow a user-defined tasks.
%%JP: should motivate this more. Mention that random sampling along does not conform to user commands, so task sampling is a way to get the model to follow commands.
First, a pool of $P$ random candidate trajectories are generated by sampling from A-MDM. The candidates are ranked according to a user-defined score function. For example, in the goal-reaching task, the score function is determined by the 2D Euclidean distance on the ground plane between the character and goal. Then the candidate trajectory with the best score is selected and its first frame is used as the character's next state. This process is repeated for every frame until the task is completed or terminated by other constraints.
%%JP: we need to rewrite this paragraph a bit. The current description is kind of confusing.

We find that the task-oriented sampling works well for simple locomotion tasks, such as target reaching (as shown in Section~\ref{target reaching}). Figure~\ref{fig_sampling_reaching} illustrates trajectories of sampled motion for this task. Compared to Motion VAE~\cite{ling2020character}, our method can often successfully navigate to the target motion directly, rather than producing less efficient circling motions. However, for more precision-oriented tasks such as joystick and path following, task-oriented sampling can still struggle to closely follow the specified commands.
\begin{figure}[t!]
    \centering
    \includegraphics[width=0.5\textwidth]{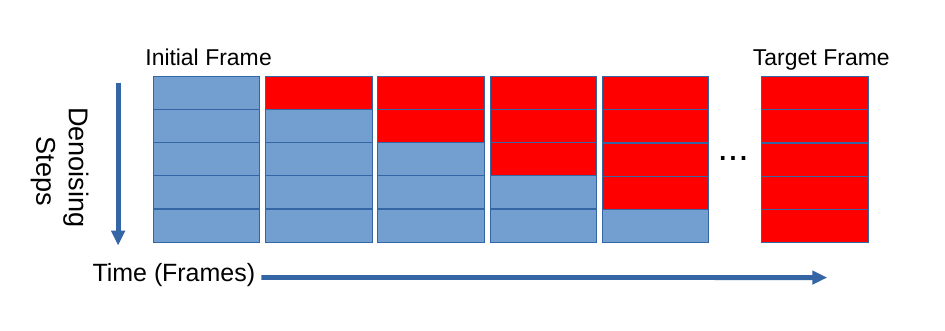}\\
    \vspace{-0.4cm}
    \caption{\textbf{Transition In-betweening through inpainting.} To generate smoother transitions, we initialize the denoising process using the target frame at different denoising steps. As the frames approach the time of the target frame, the denoising process is initialized at a later and later denoising step with the target frame, which leads the generated frames to more closely conform to the target frame.}
    \label{fig:inbw}
\end{figure}
\subsection{Synthesis via Conditional Inpainting}~\label{Method_conditional_inpainting}

\begin{figure}[h!]
    \centering
    \includegraphics[width=0.48\textwidth]{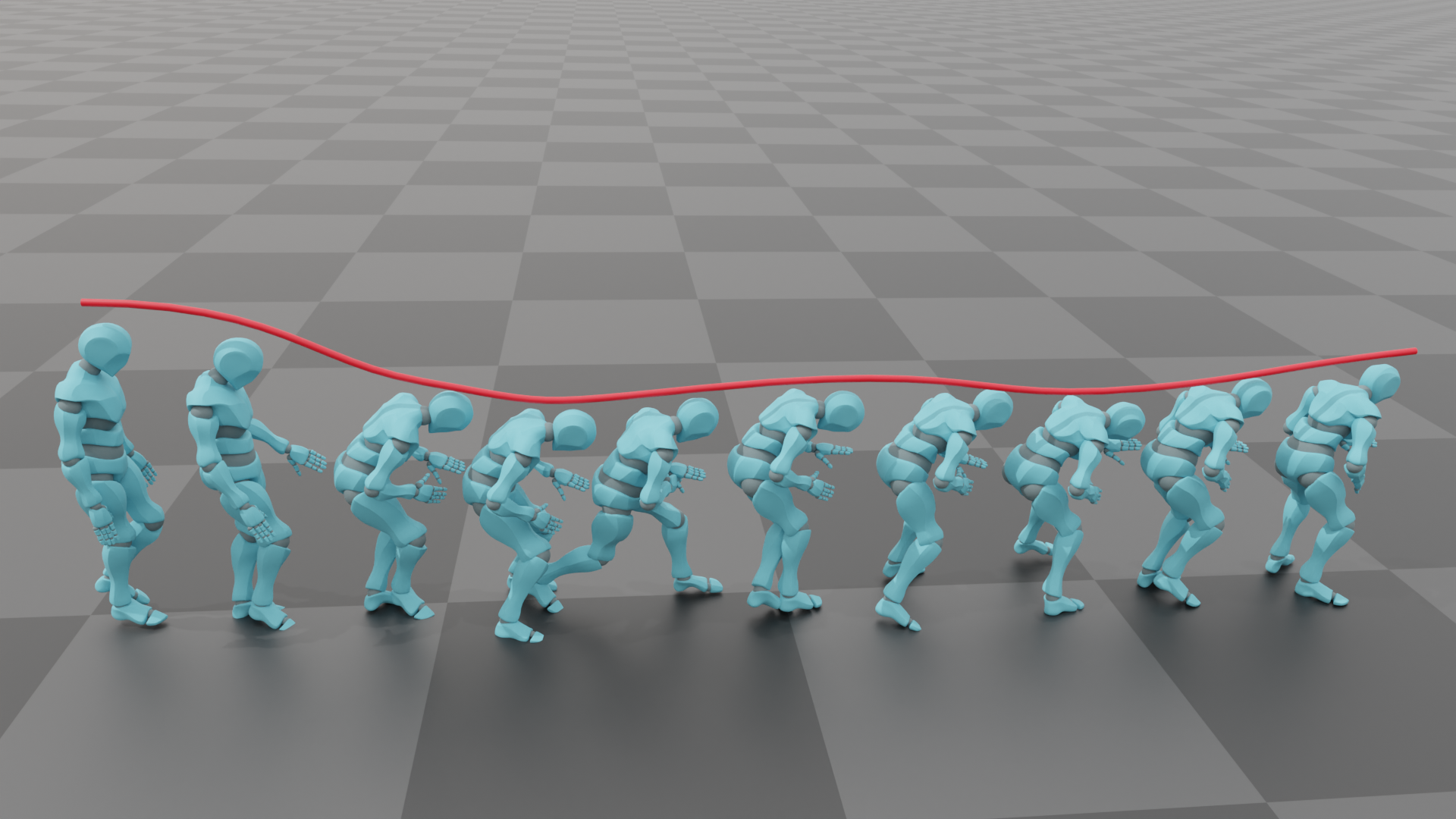}\\
    \vspace{-0.2cm}
    \caption{\textbf{Joint position control through inpainting.} Given desired trajectories for the head and root, A-MDM is able to synthesize high-quality motions via inpainting techniques without additional finetuning.}
    \label{fig:conditional}
\end{figure}

For tasks that require more precise control of the generated motions, task-oriented sampling may not be able to closely follow a user's specifications. To enable more fine-grain control, we show that A-MDM is also amenable to diffusion model-based inpainting, as shown in Figure \ref{fig:inpt}. This approach enables A-MDM to be used for a wide range of tasks without requiring further fine-tuning of the model. When applying inpainting to A-MDM, the user first specifies desired features $\tilde{x}_f$ for the motion at each frame $f$, along with a binary mask $m_f$ that records which features are explicitly specified by the user. Each entry in $m_f$ is assigned a value of 1 if the feature is specified by the user, and 0 if the feature was not assigned specific values by the user. To incorporate the user's control into the denoising process, the values of features $x_f^t$ after each denoising step is directly replaced with the values specified by the user,
\begin{align}
    \hat{x}_f^t = (1- m_f) \odot x_f^t + m_f \odot \tilde{x}_f^t, \quad x_f^t \sim q\left(x_f^t \middle| x_f^{t-1}\right),
\end{align}
where $\odot$ denotes a component-wise multiplication. The resulting frame $\hat{x}_f^t$ is then used as the input to the next denoising step. This simple inpainting method then enables A-MDM to produce coherent full body motions that precisely follow controls from the user as shown in Figure \ref{fig:conditional}. For example, the user can specify trajectories for various joints, such as the root, and A-MDM can then generate a plausible full body motion that follows the desired trajectory. We refer to this form of inpainting as \emph{spatial inpainting}.

In addition to spatial inpainting, where a full-body motion is generated from a user-specified trajectory for a subset of the joints, A-MDM can also be used for \emph{temporal inpainting}, more commonly referred to as \emph{keyframe inbetweening}. For keyframe inbetweening, the user specifies target key frames $\tilde{x}_f$ at a sparse number of frames, where each keyframe specifies values for the full state of the character. The goal then is to generate a natural motion for the intermediate frames between two adjacent keyframes. We propose a simple method to generate natural transitions between two keyframes using A-MDM by simply initializing the denoising process with the target keyframe starting at different denoising steps. Given an initial frame $x_0$ and a target keyframe $\tilde{x}_N$ at frame $N$, the target keyframe is gradually introduced into the denoising process by first initializing the denoising process at frame $f$ with the target keyframe $\tilde{x}_N$ an then applying the denoising process starting at diffusion step $t_0$,
\begin{align}
    t_0 = \left(1 - \frac{f}{N}\right) t_\mathrm{max},
\end{align}
where $t_\mathrm{max}$ is the maximum number of diffusion steps for each denoising process. For early frames $f \approx 0$ in the inbetweening process, $\tilde{x}_N$ is used to initialize the early denoising steps $t \approx t_\mathrm{max}$, and many denoising steps are then applied, which allows the model flexibility to deviate from the keyframe as needed to generate a more natural motion at the early frames. As the frames get closer to the keyframe at $f=N$, fewer and fewer denoising steps are applied to $\tilde{x}_N$, which enforces that the later frames closely match the desired keyframe, as shown in Figure \ref{fig:inbw}. In our experiments, we show that this simple inbetweening method allows A-MDM to produce realistic motions between keyframes, as well as generate natural transitions between disparate motion clips.  An example of the motion inbetweening is shown in Figure \ref{fig:inbet}, and more examples are available in the supplementary video.

%\subsection{High-Level Controller}\label{Method_sub_CbE}

\begin{figure}[h!]
    \centering
    \includegraphics[width=0.48\textwidth]{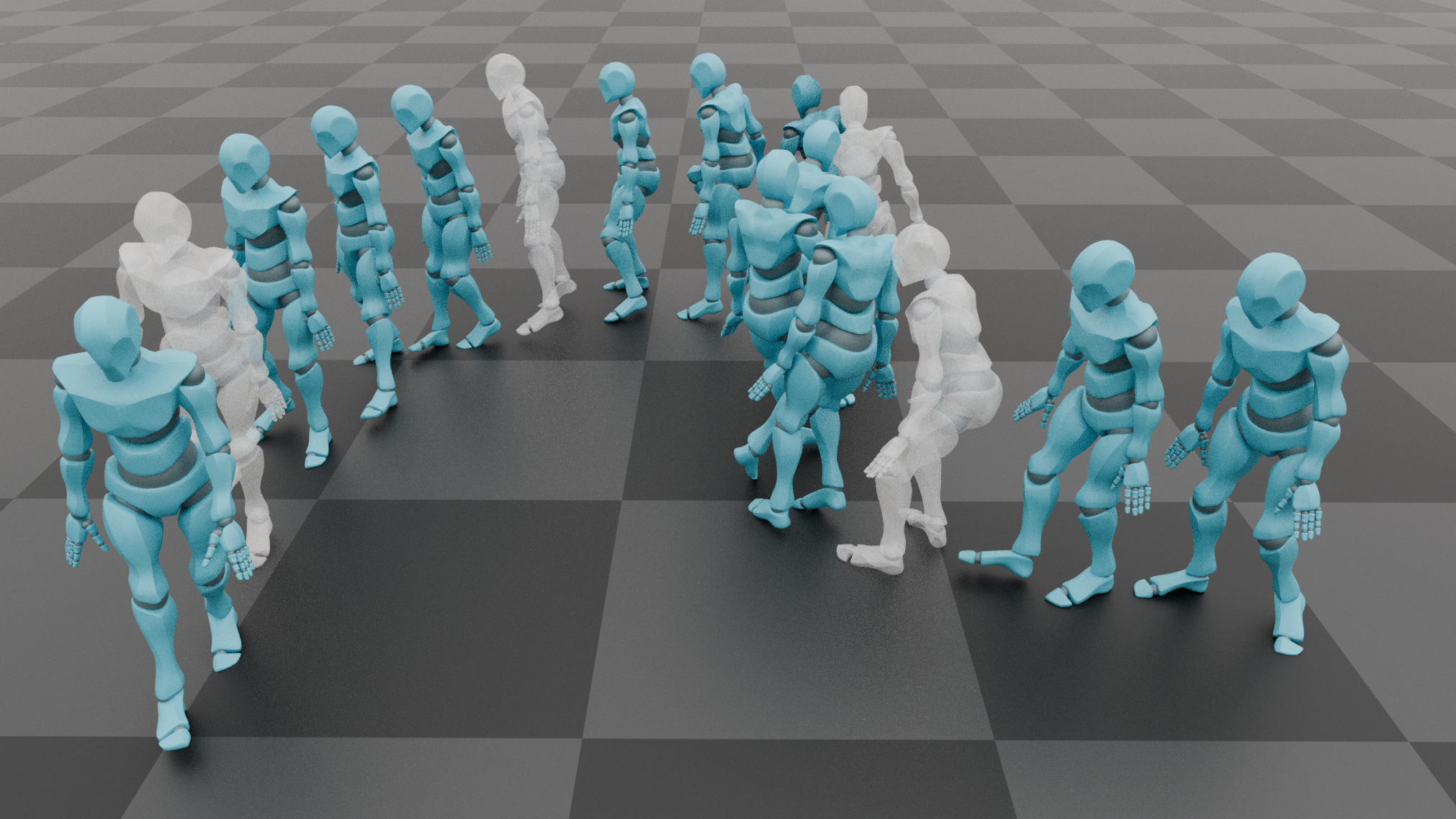}\\
    \vspace{-0.2cm}
    \caption{A-MDM can be used for key-frame in-betweening to generate plausible motions (blue) between user specified key-frames (white).}
    \label{fig:inbet}
\end{figure}

\section{Hierarchical Control}\label{Method_Controll}
To address the limitation of task-oriented sampling and inpainting, we propose using hierarchical reinforcement learning to train high-level controllers that steer the base A-MDM model towards producing motions that follow the user's commands.

Once the high-level controller is trained, the proposed approach demonstrates significantly more effective performance on new tasks compared to the sampling-based techniques. Furthermore, we show that this approach offers greater flexibility when completing the same tasks than conditional inpainting, as shown in Figure \ref{trajetory_control_vs_inpainting}.

\subsection{Reinforcement Learning}\label{Method_sub_RL}
Reinforcement Learning (DRL) is used to train task-specific policies for directing the low-level A-MDM model towards completing new tasks. With reinforcement learning, policies are trained by optimizing a policy's expected return:

\begin{equation}\label{e4_rl}
J_{RL}(\pi)=\mathbb{E}_{\tau \sim p(\tau | \pi)}\left[\sum_{f=0}^\infty{\gamma^fr(s_f,a_f)} \right],
\end{equation}

where $p(\tau | \pi)$ is the distribution of trajectories $\tau$ induced by a policy $\pi$, $r(s_f,a_f)$ is the reward that defines a desired task, and $\gamma \in [0,1]$ is a discount factor. The agent receives a reward at each timestep $f$ by executing an action $a_f$ at state $s_f$. We optimize this objective using Proximal Policy Optimization (PPO) \cite{schulman2017proximal} to train task-specific high-level controllers for each task. Noted we use $f$ to represent the MDP timestep to distinguish it from the diffusion timestep $t$. At each frame $f$, the agent takes as input the observation of the current state $s_f$ of the environment, and then selects an action that is used to steer the sampling procedure of the pretrained A-MDM.

\subsection{Hierarchical Control for A-MDM}
\begin{figure}[t!]\label{hier_control}
    \centering
    \includegraphics[width=0.5\textwidth]{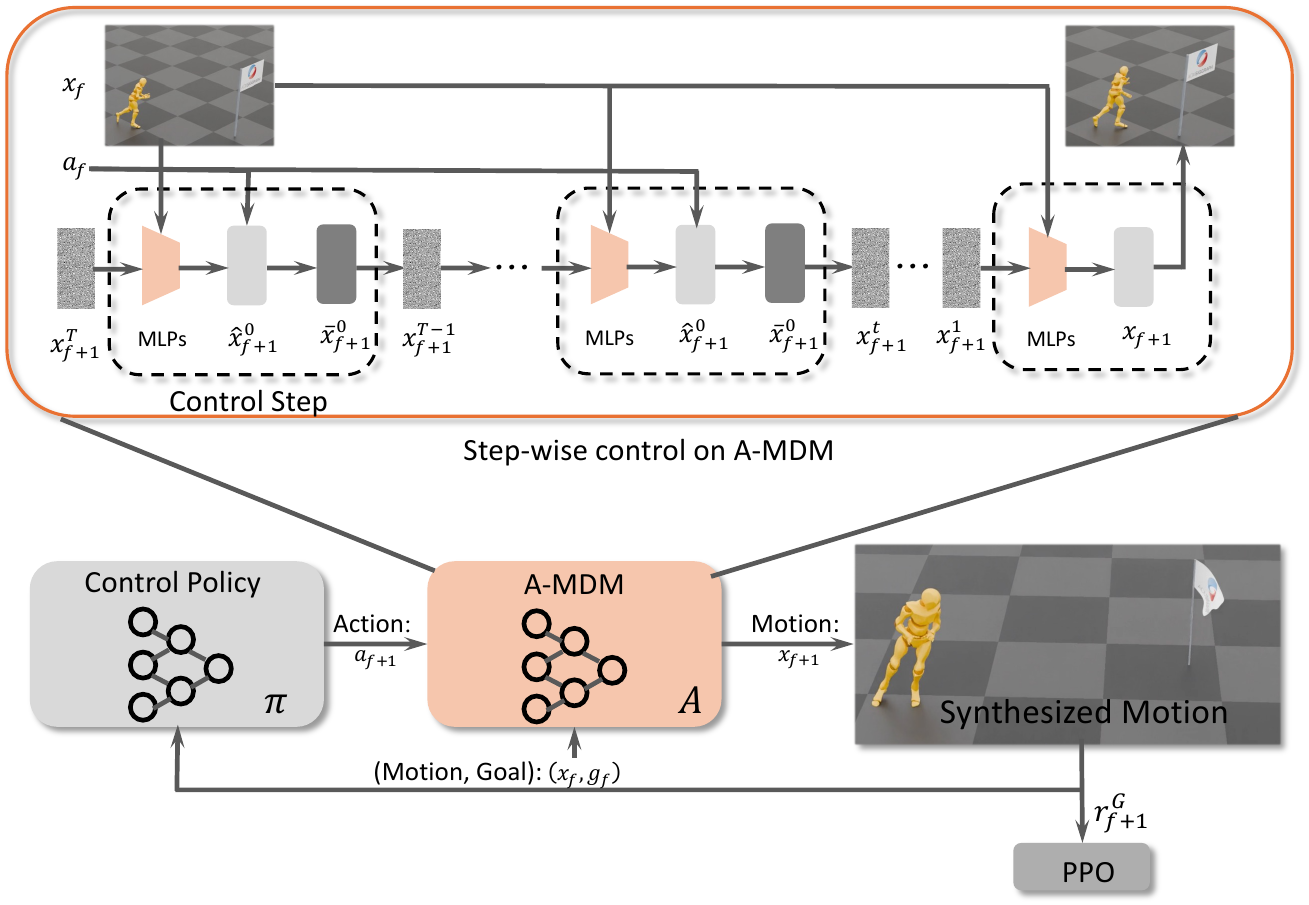}\\
    \vspace{-0.2cm}
    \caption{Hierarchical Control for A-MDM. The high-level controller predicts the residual vectors $a_{f}^t$ of $\hat{x}_f^t$ to steer the denoising process of the base A-MDM.}
    \label{f2_rl2}
\end{figure}

In previous VAE-based models \citep{ling2020character,yao2022cvae,won2022physics}, a base VAE model is trained to produce a latent space that can be directly used to sample motions for the next frame. Hierarchical controllers are then trained to sample latent codes from the latent manifold of the base VAE model. In contrast, it's not straightforward to train a hierarchical controller to control diffusion model, where Gaussian noises is introduced iteratively during inference. To apply a pre-trained A-MDM model to new tasks, we train task-specific high-level controllers to steer the denoising process to produce motions that satisfy desired task objectives. Figure~\ref{f2_rl2} provides a schematic illustration of our hierarchical controller. Given the character's current state $x_{f}$, the high-level controller predicts an action $a_{f} = \{a_{f}^T,..,a_{f}^1\}$, which consists of multiple residual vectors for different denoising steps. Each residual vector $a_{f}^t$ is applied as a perturbation to the output of a denoising step $t$,
\begin{align}
\bar{x}_{f+1}^t = \hat{x}_{f+1}^t + {w^t}a_f^t.
\end{align}
The perturbed output $\bar{x}_{f+1}^t$ is then used as the input for the next denoising step $t-1$. $w^t$ is a weight that anneals the perturbation according to the denoising step. The annealing schedule is designed to gradually attenuate towards the end of the denoising steps to prevent sudden changes. This annealing is crucial since large perturbations from the high-level controller at the later denoising steps can lead to motion artifacts.

\section{Tasks}~\label{Method_loco_motion}

To evaluate the effectiveness of our hierarchical model, we evaluate our framework through three tasks: Target Reaching (Section~\ref{target reaching}), Joystick Control (Section~\ref{joystick}), and Path Following (Section~\ref{path following}).

\subsection{Target Reaching}~\label{target reaching}
Target reaching is a locomotion task in which the character is required to move towards a user-designated target location. The location of the target can be altered by the user at any time within an episode.

The task is completed if the character's hip joint is within 15~$cm$ of the target. Once the character has reached a target, a new target is randomly placed within a 20m$\times$20m region around the character's current location. 
The task observation of the high-level controller consists of the target position $G^*$ relative to the character. 

The reward is computed using a combination of a distance reward between the character's position and $G^*$, as well as the progress made towards $G^*$. When the agent reaches the goal, the agent will also receive a one-time bonus reward before a new target is set.

\subsection{Joystick Control}~\label{joystick}
The Joystick Control task allows the user to specify the speed and orientation of the character.
The character is trained to generate natural motions that follow the user specified velocity and heading. To train the high-level controller for this task, random joystick commands are generated by changing the desired direction and speed every 120 and 240 frames respectively. The task observation consists of the facing direction $d_r^g$, which is uniformly sampled between 0 and 2$\pi$, as well as the desired velocity $d_v^g$, which is uniformly sampled from 0.6m/s to 7.2m/s. The reward function is given by
\begin{equation}
    r = e^{cos(d_r - d_r^g) - 1} \times e^{-|d_v - d_v^g|},
\end{equation}
where $d_r$ is the character's angular velocity, and $d_v$ is the character's planar linear velocity. To ensure that the policy satisfies both of these objectives simultaneously, the two rewards are combined multiplicatively.

\subsection{Path Following}~\label{path following}
In the Path Following task, the character is required to navigate along a user-defined path composed of several waypoints. The character must reach each waypoint in the specified order. The task observation consists of the locations of the future waypoints along the pre-defined path. During training, paths are procedurally generated by randomly sampling linear and angular accelerations for each waypoint along the path. Once trained, the controller is able to follow arbitrary paths at runtime. Figure \ref{trajetory_control_vs_inpainting} shows examples of our hierarchical controller following user-specified trajectories. The base A-MDM model enables the character to follow the target trajectories with natural behaviors.

\section{Experiments}

\begin{table*}[ht]
\caption{Comparisons on AMASS, 100STYLE and LaFAN1. 50 motion sequences are generated starting at fixed initial states. Each motion is 60 frames long when evaluating ADE and 150 frames long for calculating APD.  (unit: cm). }\label{regular_eval}
\centering
  {\begin{tabular}{l cccccccc}
    \toprule
    & APD $\uparrow$ & ADE $\downarrow$ & FDE $\downarrow$ & FS $\downarrow$  & Bone.Err $\downarrow$  & Jnt.Accel$\rightarrow$ & Pen.Freq $\downarrow$ & Pen.Dist $\downarrow$\\
    \midrule
    \textbf{AMASS} & - & -& - & - & - & 7.28  & - & -\\ 
    \hline 
    
      MVAE   &  40.97$\pm$3.97 & 24.42$\pm$1.04  &  42.22$\pm$3.64 & 1.44$\pm$0.12 & 0.98$\pm$0.12 & 7.54 $\pm$0.51 & 1.94$\pm$0.56{\%} & 1.26$\pm$0.82 \\ 
      HuMoR   & 44.95$\pm$4.49 & 17.96$\pm$1.17  &  41.10$\pm$3.98 & 1.35$\pm$0.11 & 1.04$\pm$0.07 & 7.69$\pm$0.49 & 1.78$\pm$0.62{\%} &1.17$\pm$0.93  \\
      AMDM (Ours)  & \textbf{61.08}$\pm$1.35  & \textbf{10.40}$\pm$0.66 & \textbf{21.12}$\pm$1.16 & \textbf{1.06}$\pm$0.11 & \textbf{0.82}$\pm$0.04& \textbf{7.26}$\pm$0.19 & \textbf{0.4}$\pm$0.03{\%} & \textbf{1.07}$\pm$0.32  \\
    \hline 
    \textbf{100STYLE}  & - & -& - & - & - & 9.87  & - & - \\
    \hline 
      MVAE   &  58.44$\pm$1.36 & 23.47$\pm$0.42  & 48.24$\pm$1.52  & 1.62$\pm$0.05 & 0.25$\pm$0.02  & 9.48$\pm$0.52 & 1.87$\pm$0.84{\%} & 0.06$\pm$0.02 \\
      HuMoR   & 68.25$\pm$1.45  & 17.41$\pm$0.47 & 43.34$\pm$1.74  &  1.57$\pm$0.06 &  0.23$\pm$0.02 & \textbf{9.59}$\pm$0.48 & 1.80$\pm$0.92{\%} & 0.07$\pm$0.04\\
      AMDM (Ours)  & \textbf{102.52}$\pm$1.17 & \textbf{10.36}$\pm$0.22 & \textbf{24.58}$\pm$0.06  & \textbf{1.53}$\pm$0.02  & \textbf{0.19}$\pm$0.01& 9.37$\pm$0.28 & \textbf{1.56}$\pm$0.24  & \textbf{0.04}$\pm$0.01\\
    \hline 
    \textbf{LaFAN1} &-&-&-& - & -& 12.56 &-& -\\
    \hline 
      MVAE  & 110.48 $\pm$5.67 & 35.50$\pm$3.43  & 82.93$\pm$4.8  & 2.42$\pm$0.49 & 0.66$\pm$0.02  & 12.81$\pm$0.37  & 0.83$\pm$0.09$\%$ & 0.65$\pm$0.14\\
      HuMoR & 132.76$\pm$4.25  & 24.35$\pm$2.19 & 41.61$\pm$2.97  &  2.20$\pm$0.26 & \textbf{0.53}$\pm$0.03  & 13.03$\pm$1.19  & \textbf{0.71}$\pm$0.03$\%$ &0.50$\pm$0.06\\
      AMDM (Ours)  & \textbf{134.92}$\pm$6.01 & \textbf{14.22}$\pm$2.20 & \textbf{34.53}$\pm$4.16 & \textbf{2.10}$\pm$0.56  & 0.54$\pm$0.02 & \textbf{12.74}$\pm$0.35 & 0.76$\pm$0.07$\%$ & \textbf{0.49}$\pm$0.12  \\
    \bottomrule
  \end{tabular}}
\end{table*}
In this section, we demonstrate the effectiveness of our A-MDM framework. First, we evaluate the base A-MDM model and compare it with prior VAE-based models (\textit{e.g.}, MVAE and HuMoR ~\citep{ling2020character, rempe2021humor}) to showcase the effectiveness of our design choices for motion modeling. We then show that A-MDM can be combined with hierarchical RL to solve downstream motion control tasks.

\subsection{Experimental Setup}\label{env}
Our framework is implemented with Pytorch. The experiments with hierarchical controllers are conducted using OpenAI Gym~\cite{openaigym}. All experiments are performed on a PC with an NVIDIA GeForce 4090 GPU and Intel Core i9-13900K. 

\subsection{Dataset}\label{dataset}

Our framework is evaluated on three datasets that vary in terms of size and motion diversity:
\begin{enumerate}
    \item \textbf{100STYLE} contains more than 4,000,000 frames of motion capture data of 100 diverse styles of locomotion ~\citep{mason2022local}. For each style, the dataset consists of motions with different velocities (idle, running, and walking), and different headings (sidewalk, forward, and backward). For evaluation, we partition the data according to the official training and testing splits.
    \item \textbf{AMASS} is a large-scale motion capture database, represented with a unified SMPL body model ~\citep{mahmood2019amass}. We evaluate our model on this dataset to show that A-MDM can be effectively applied to large dataset and produce better performance than prior auto-regressive models.
    \item \textbf{LaFAN1} is a high-quality motion capture dataset containing highly dynamic motions, including dancing, crawling, and fast locomotions, with a size of more than 400,000 frames ~\citep{harvey2020robust}. We train our model with a subset of the original LaFAN1 dataset, excluding environment interaction motions.
\end{enumerate}
All motion clips are standardized to a frame rate of 30 Hz.

\begin{table*}[ht]
\caption{To evaluate the models' generalization capabilities when generating new motions not in the dataset, we use the models to generate continuation motions starting at the last frame of motion clips in the dataset. We compare the models on the AMASS, 100STYLE, and LaFAN1 datasets. (unit: cm).
}\label{continuation_eval}
\centering
  \begin{tabular}{lccccc}
    \toprule
    & APD $\uparrow$ & FS $\downarrow$  & Bone.Err $\downarrow$  & Pen.Freq$\downarrow$ & Pen.Dist$\downarrow$\\
    \hline 
    \textbf{AMASS} \\
    %\hline 
    \midrule
      MVAE   & 41.27$\pm$4.06  & 1.52$\pm$0.14  & 2.96$\pm$0.42 & 2.07$\pm$0.89{\%} & 1.14$\pm$0.73 \\ 
      HuMoR  & 43.26$\pm$5.77 & 1.47$\pm$0.16  & 3.05$\pm$0.35 & 1.89$\pm$1.07{\%}  &1.01$\pm$ 0.91  \\
      AMDM (Ours)  & \textbf{59.91}$\pm$1.03  & \textbf{1.10}$\pm$0.14 & \textbf{2.01}$\pm$0.06 & \textbf{0.38}$\pm$0.05{\%} & \textbf{0.89}$\pm$ 0.22\\
    \hline 
    \textbf{100STYLE} \\
    \hline 
      MVAE  & 59.44$\pm$0.94 & 1.66$\pm$0.05  & 0.26$\pm$0.02  & 1.78$\pm$0.85{\%} & 0.06$\pm$ 0.04  \\
      HuMoR   &67.83$\pm$0.82  & 1.64$\pm$0.05  & 0.23$\pm$0.02  & 1.86$\pm$0.95{\%} & 0.04$\pm$ 0.03  \\
      AMDM (Ours)  &\textbf{109.11}$\pm$0.07  & \textbf{1.56}$\pm$0.02 & \textbf{0.20}$\pm$0.02  & \textbf{1.70}$\pm$0.38{\%} & \textbf{0.03}$\pm$0.01\\
    \hline 
    \textbf{LaFAN1} \\
    \hline 
      MVAE  &  96.26$\pm$16.40 & 2.15$\pm$0.47  & 0.57$\pm$0.20  & 1.01$\pm$0.46$\%$ & 0.33$\pm$0.17\\
      HuMoR  & 123.78$\pm$8.64  & 2.07$\pm$0.05 & \textbf{0.43}$\pm$0.05  & 1.07$\pm$0.24$\%$ &0.46$\pm$0.21 \\
      AMDM (Ours)  & \textbf{124.46}$\pm$12.69 & \textbf{1.94}$\pm$0.01  & 0.45$\pm$0.16 & \textbf{0.86}$\pm$0.08$\%$ &\textbf{0.30}$\pm$0.03  \\
    \bottomrule
  \end{tabular}
\end{table*}

\subsection{Evaluation Metrics}\label{metrics}
 To evaluate the quality of the generated motion, we generate 50 motion sequences from each model, where each motion is initialized in fixed initial states. Each motion has a length of 150 frames. To quantify the diversity among these generated motions, we measure the Average Pairwise Distance (APD): 
\begin{equation}
%APD(X) = \frac{1}{N(N-1)} \sum_{j=1}^{J}\sum_{i=1}^{N} ||B_{ji}-A_{ji}||_2, 
\mathrm{APD}(x_i...x_K) = \frac{1}{K(K-1)}\sum_{i=1}^{K}\sum_{j\neq i}^{K}||x_{i}-x_{j}||, 
\end{equation}
where $K$ is the total number of sequences, and $x_i$ denotes one of the generated motion. A large APD indicates the model can generate diverse motions. However, a limitation of this metric is its tendency to favour models that generate motions with faster velocities, which are more likely to lead to larger joint position differences. 

To evaluate the similarity of the generated motion with respect to the original motion data, we calculated the Average Displacement Error (ADE) on the first 60 frames of the same 50 generated motions:
\begin{equation}
\mathrm{ADE}(x_i...x_K, x_{ref}) = {\min_{i\in{K}}}||x_{i}-x_{ref}||,
\end{equation}
where $K$ is the total number of clips, and $x_i$ is the $i$-th clip of the generated motions, $x_{ref}$ represents the ground truth motion that shared the initial state with the generated motions. ADE quantifies the distance between the distribution of generated motions and the motion data.  

To evaluate how close the endpoint of the predicted motion is to endpoint of the ground-truth motion, we calculate the Final Displacement Error (FDE). FDE measures the distance between the joint positions of the generated motion to the ground-truth motion with a prediction horizon of 60 frames. 50 different motions are generated using A-MDM, and the FDE is recorded using the closest sample from the ground-truth trajectory. FDE provides an additional evaluation metric to assess the similiarity between the distribution of generated motions and the ground truth data.

However, APD and ADE alone are not sufficient for a comprehensive evaluation of motion quality. A model can produce a motion sequence with notable artifacts, such as deformation, and still achieve high APD and acceptable ADE scores. To address this issue, we leverage the assumption that the character's bones are rigid bodies. This allows us to gauge the extent of deformation by comparing the bone lengths in generated motions against those in the ground truth mocap data. Specifically, we calculate the average deviation of bone lengths in the generated joint positions from those of a standard skeleton. We refer to this measure as the ``Bone Length Error''. Through empirical analysis, we have found that this metric is well correlated with motion quality and robustness of the motion synthesis model. Bone Length Error is computed according to:
\begin{equation}
\mathrm{Bone\_Err} = \frac{1}{B}\sum_{b\in{B}} ||\hat{L}_{b}-L_{b}||,
\end{equation} 
where $B$ represents the number of bones in the character's body. $\hat{L_b}$ denotes the length of a bone estimated from the joint positions of the generated motions, and $L_b$ is the ground truth bone length from the dataset. 

We additionally record the occurrence of common artifacts in motion synthesis, such as foot penetration frequency (Pen.Freq), average foot penetration distance (Pen.Dist), foot sliding (FS), and joint acceleration (Jnt.Accel). We follow the implementation of \citet{ling2020character} to evaluate foot sliding. For evaluating foot penetration, we adopt the methodology outlined from \citet{rempe2021humor}.
\subsubsection{Random Sampling}  
In this experiment, we benchmark A-MDM against previous auto-regressive motion models in the context of random synthesis, as outlined in Section \ref{metrics}. This comparison is conducted using AMASS, 100STYLE, and our subset of the LaFAN1 datasets. For the evaluation presented in Table \ref{regular_eval}, initial states are uniformly sampled from each dataset. To evaluate the models' ability to generalize and generate motions not in the original dataset, we introduce an evaluation scheme, termed ``Motion Continuation'', which assesses a model's ability to extend a ground truth motion sequence. In this setting, initial states are extracted from the last 30 frames of the mocap clips, and the model then generates a motion that extends beyond the end of the original motion clip. The results of this assessment are documented in Table \ref{continuation_eval}. In general, A-MDM is capable of generating motions with greater diversity and higher fidelity compared to other auto-regressive generative models. 

\begin{figure*}[t!]
   {\includegraphics[width=18cm]{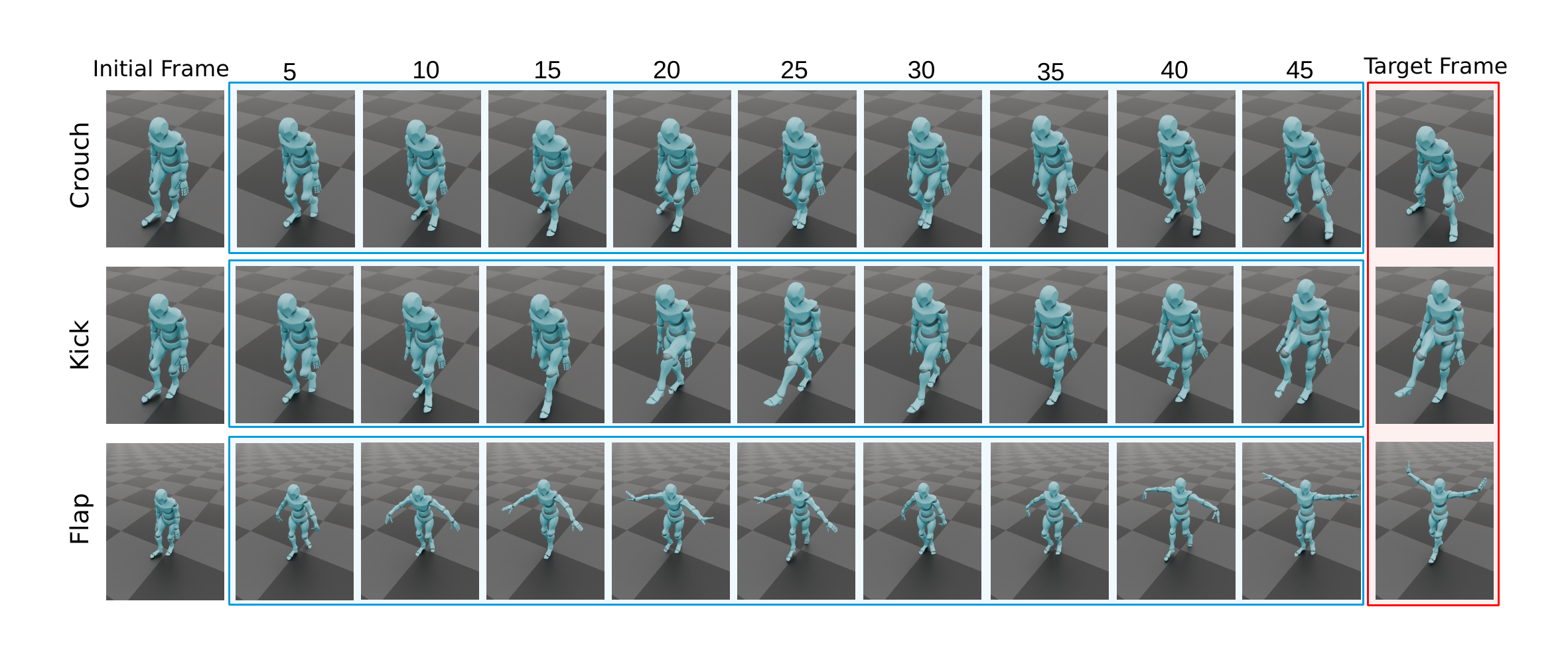}}\\
   \vspace{-0.7cm}
    \caption{\textbf{Qualitative results for Motion In-betweening via inpainting}. Red represents the target frame, and blue indicates the inbtweening frames. We use a 40-step A-MDM to generate the transition between the same initial frame and different target frames. A-MDM is able to generate natural transitions between keyframes in real-time.}
    \label{inbetween_vis}
\end{figure*}

\subsubsection{Denoising Steps and Inference Time}
Next, we investigate the influence of different diffusion steps during inference. The performance of A-MDM trained with different number of denoising steps are summarized in Tables \ref{step_eval} and \ref{step_eval_lafan1}. We observe that models trained on larger datasets tend to require a greater number of steps to reach optimal performance. Given A-MDM's primary objective is for real-time motion synthesis, it is crucial to strike a balance between computational efficiency and motion quality. Through our experiments on 100STYLE, we find that 40 denoising steps offers an effective trade-off between reducing motion artifacts and preserving motion diversity. As result, this 40-step configuration is used in subsequent experiments.   
\begin{table}[h!]
\caption{Comparison of A-MDM with different number of diffusion steps on 100STYLE. (unit:cm) }\label{step_eval}
  \begin{tabular}{l cccccc}
    \toprule
     Step & APD $\uparrow$ & ADE $\downarrow$ & Time(s) $\downarrow$ & FS $\downarrow$\\ 
    \midrule
      10 & 96.70 & 10.44 &  \textbf{0.009} & 1.55 \\
      20 & 101.17 & 10.16 & 0.012 & 1.57 \\
      30 & 101.19 & \textbf{10.31} & 0.015 & 1.55 \\
      \textbf{40} & \textbf{102.52} & 10.36 & 0.021 & \textbf{1.53} \\       50 & 100.97 & 10.36 & 0.026  & \textbf{1.53} \\
    \bottomrule
  \end{tabular}
\end{table}

\begin{table}[h!]
\caption{Comparison of A-MDM with different numbers of diffusion steps on the \textbf{full} LaFAN1, excluding environment interaction motions. Distance error units are measured in cm.}\label{step_eval_lafan1}
  \begin{tabular}{l ccccccc}
    \toprule
     Step & APD $\uparrow$ & ADE  $\downarrow$ & Time(ms) $\downarrow$ & FS $\downarrow$ &\\ 
    \midrule
      1 & 53.12 & 25.05 & \textbf{0.55} & \textbf{1.60} \\
      2 & 100.43 & 19.00 & 1.08 & 1.72  \\
      5 & 111.78 & 17.96 & 2.80 & 2.06 \\
      10 & 118.93 & 16.91 & 5.31 & 2.02 \\
      \textbf{25} & 130.34 & 18.66 & 12.95 & 2.11\\
      40 & 128.91 & 18.01 & 20.96 & 2.22 \\
      50 & 129.32 & 18.11 & 26.28 & 2.15  \\
      100 & \textbf{130.28} & \textbf{16.18} & 52.78 & 2.06 \\
      
    \bottomrule
  \end{tabular}
\end{table}

Our findings show that the models with the fewest denoising steps exhibits the lowest motion diversity, yet they tend to exhibit less foot sliding. Fewer denoising steps are more susceptible to mode-collapse, leading to models that predominantly generates simpler motions like standing and walking, which are less prone to producing artifacts such as bone length distortion or foot sliding. Please refer to our supplementary video for a qualitative comparison of models with different numbers of denoising steps.

With a larger number of denoising steps, the model exhibits a higher Average Pairwise Distance (APD), indicating an increase in the model's capability to synthesize diverse and complex motions. However, the improvements from increasing the number of denoising steps begin to attenuate, and inference speed takes precedence over further increases in denoising steps.

\subsubsection{Task Oriented Sampling}

\begin{figure}[!t]
     \hfill
    \subfigure[HuMoR]{\includegraphics[width=4cm]{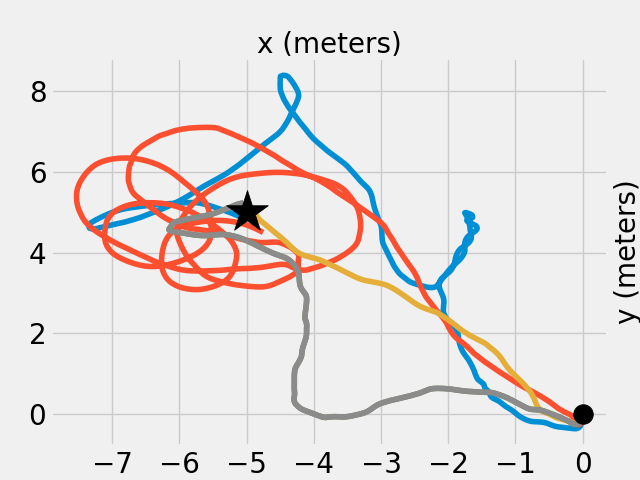}}
    \hfill
    \subfigure[A-MDM]{\includegraphics[width=4cm]{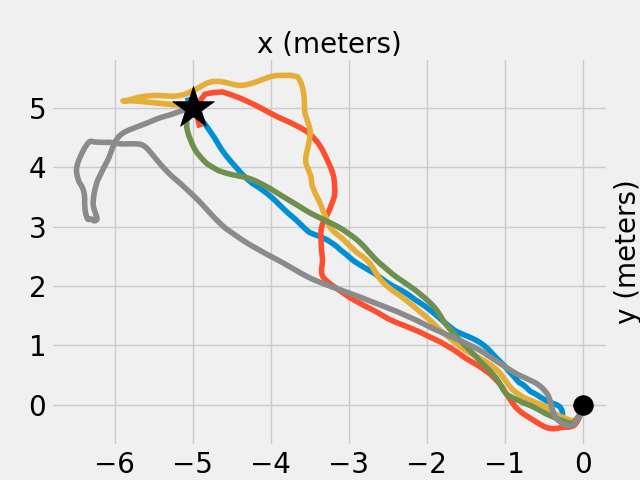}}
    \hfill
    \caption{Task-oriented sampling using HuMoR (\textbf{Left}) vs. A-MDM (\textbf{Right}). The trajectories of A-MDM are more direct and take fewer steps. Results are generated using models trained on 100STYLE.}\label{fig_sampling_reaching}
  \end{figure}
  
In this section, we present a comparison of the performance of the VAE-based models and A-MDM on the target-reaching task using task-oriented sampling. We conducted five trials for each base model, as depicted in different colors in Figure \ref{fig_sampling_reaching}. 

At each step, the base model generates 100 candidate successor states through random sampling, and the candidate state with the shortest distance to the target is selected as the next frame.

For this experiment, we initialize the character in a fixed pose at the origin (black dots) and set the target at position (-5m, 5m)
(black stars) as show in Figure~\ref{fig_sampling_reaching}. We compare A-MDM with a HuMor model trained on the 100STYLE dataset~\citep{mason2022local}.  
In Figure~\ref{fig_sampling_reaching}, the trajectories generated by HuMor are consistent with the findings from \citet{ling2020character}, where VAE-based methods tend to wander instead of taking the most direct path to the target, especially when they are close to the target. In contrast, A-MDM is able to follow more direct paths toward the target. This improvement may be attributed to more diverse motions when sampling from A-MDM, thereby providing the character with more flexible options, leading to significantly faster task completion. However, due to its inherent short-sightedness, Task Oriented Sampling often fails to identify the optimal behavior for completing a task. While task-oriented sampling is a useful tool for evaluating auto-regressive motion synthesis models, it may not always yield the best results in practice. In the upcoming experiments, we will explore strategies to leverage the base A-MDM model more effectively to further enhance performance on downstream tasks.

\begin{figure}[t!]
    %I will draw them on a single plot
    \subfigure[LaFAN1]{\includegraphics[width=4cm]{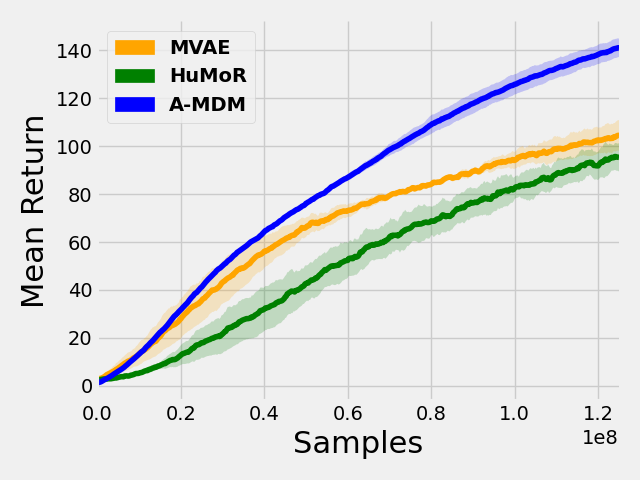}}
    \subfigure[100STYLE]{\includegraphics[width=4cm]{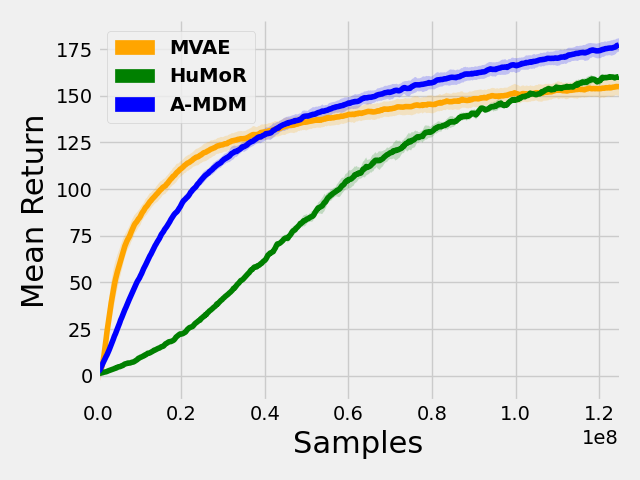}}
    
    \caption{Learning Curve of Target Reaching Hierarchical Controllers. A-MDM achieves higher returns on new tasks compared to the other hierarchical models.}
    \label{curve}
\end{figure}

\subsection{Hierarchical Control}\label{exp:control}
\begin{figure}[t!]
    \hfill
    \subfigure[Target set 1 (short distance)]{\includegraphics[width=4cm]{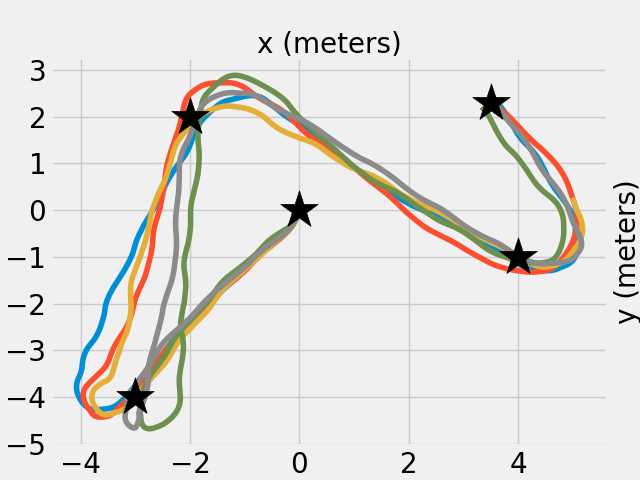}}
    \hfill
    \subfigure[Target set 2 (longer distance)]{\includegraphics[width=4cm]{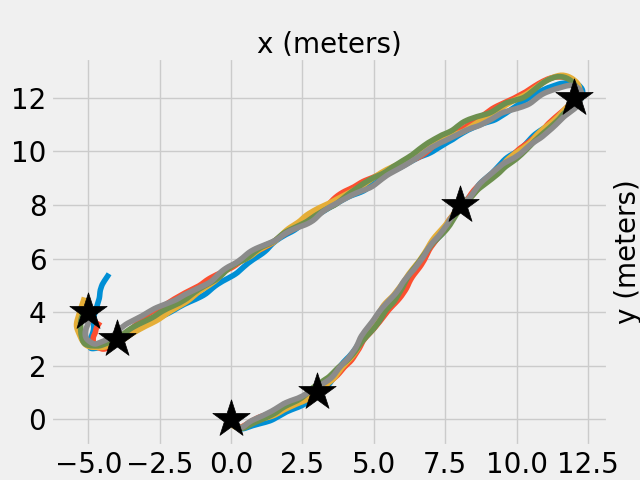}}
    \hfill
    %probably I should draw them on the same plot
    \caption{\textbf{Trajectories for Target Reaching with Hierarchical Control.} The stars stand for a fixed set of targets and lines of various colors represent different runs starting from the same initial character state. We show that our hierarchical model is capable of generating diverse motions, with the same initial state and target goals. }
    \label{rl_target}
\end{figure}

To evaluate the effectiveness of hierarchical control using different auto-regressive models, we compare the learning curve when training high-level controllers for new tasks. The base model for each method are trained on the same dataset. For the MVAE, the hierarchical controller's training adheres to \citet{ling2020character}. In the case of HuMoR, the hierarchical controller is trained to predict the residual latent of the output of the prior network. For A-MDM, the training of its hierarchical controller follows the procedures detailed in the Section \ref{Method_Controll}. Figure \ref{curve} compares the learning curves of the various models. A-MDM is able to achieve consistently higher returns compared to the other methods.

\begin{figure}[!t]
    \subfigure[Comparison between Inpainting and Hierarchical Control in trajectory following]{\includegraphics[width=8.5cm]{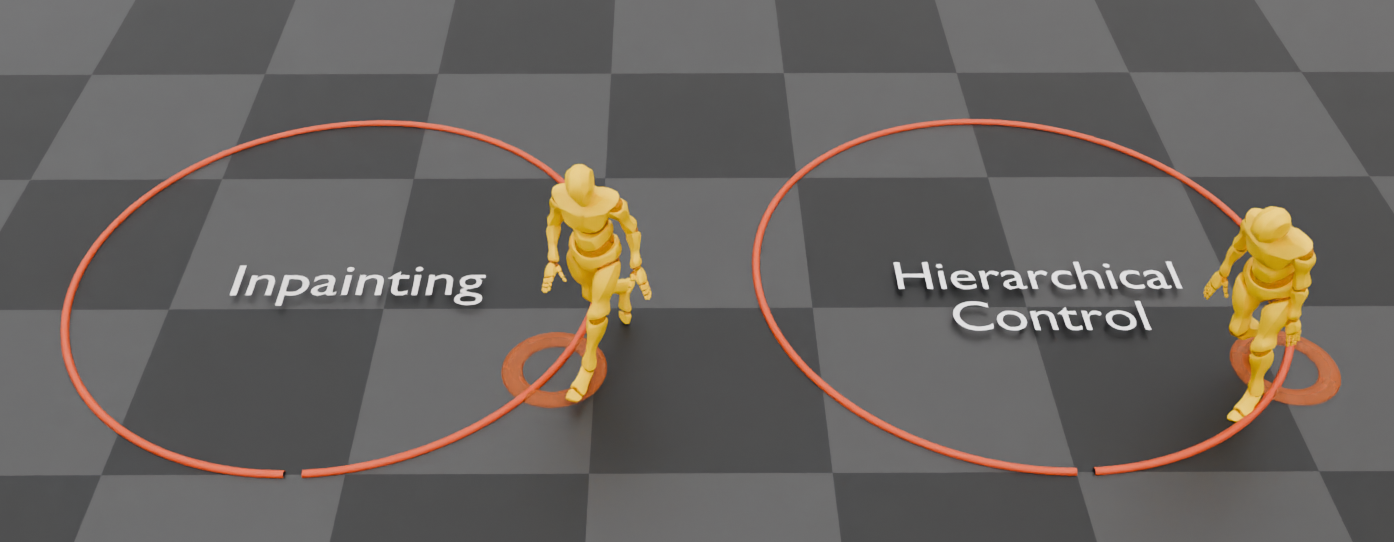}}
    \vfill
    \subfigure[Inpainting]{\includegraphics[width=4cm]{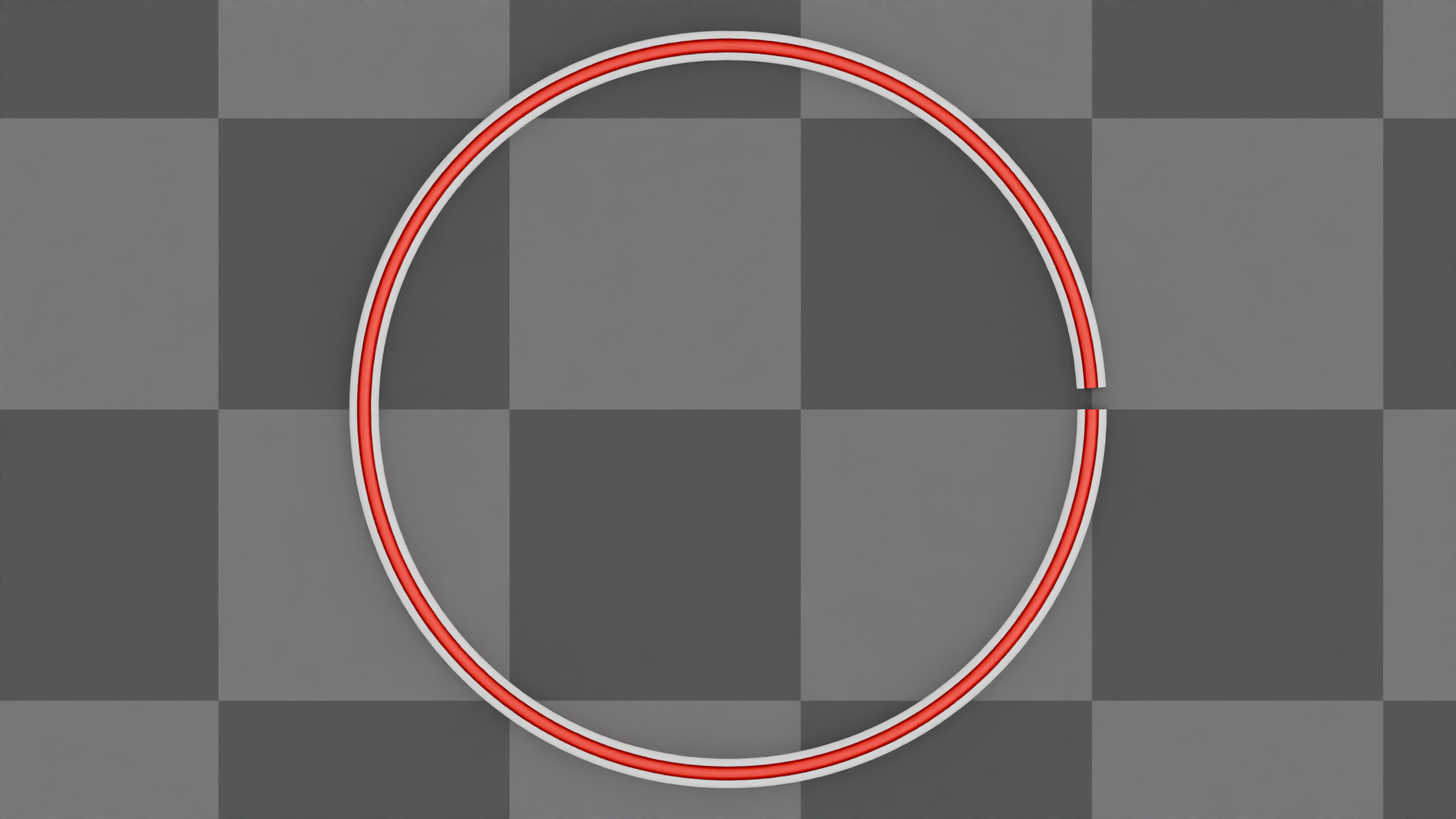}}
    \hfill
    \subfigure[Hierarchical Control]{\includegraphics[width=4cm]{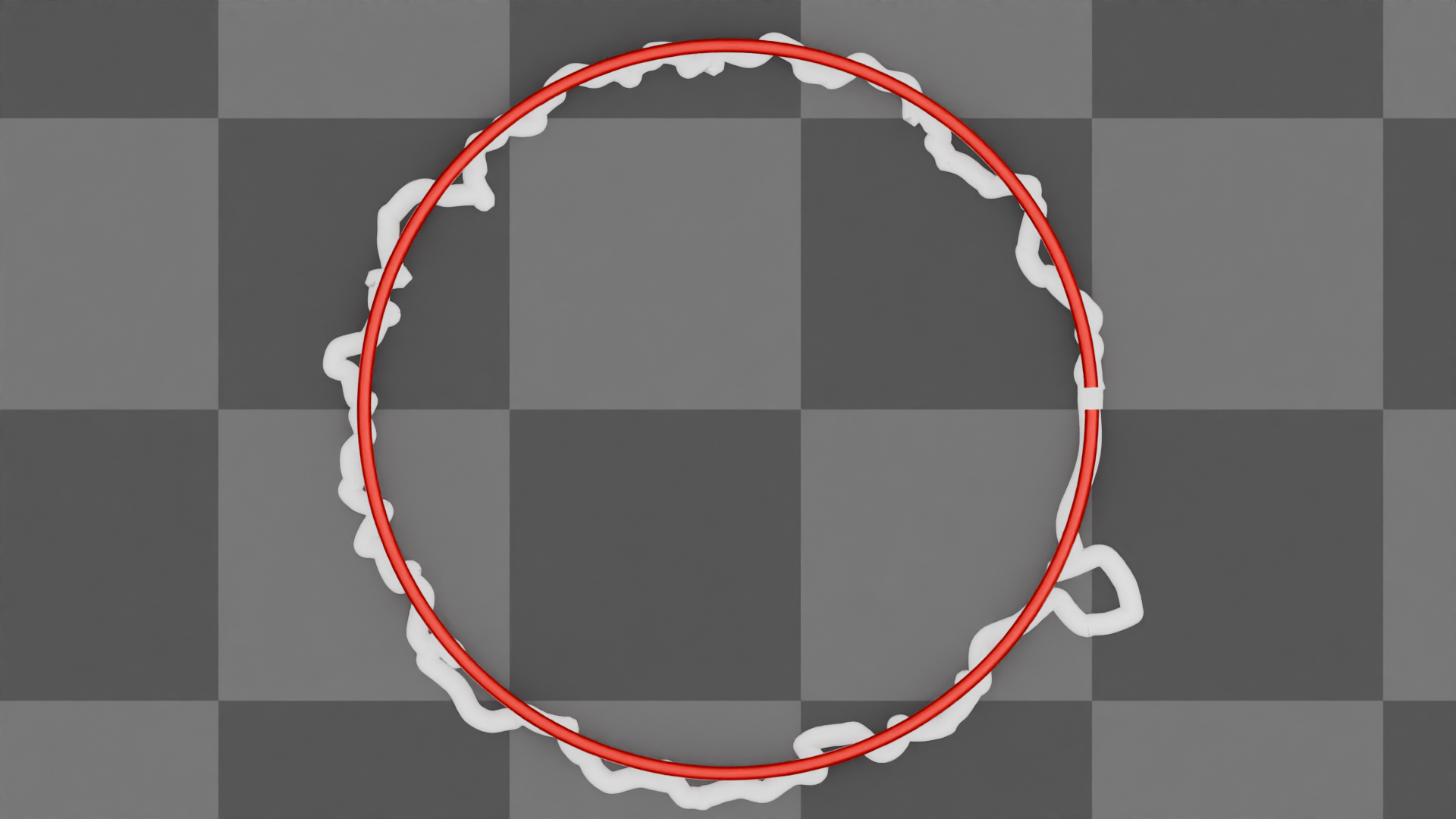}}

    \caption{Character trajectory (in white) from inpainting (\textbf{b}) and hierarchical control(\textbf{c}) when following a user-specified circular trajectory (in red). The trajectory of inpainting matches user's target trajectory exactly, while the hierarchical controller can deviate from the target trajectory as needed in order to produce more natural motions.}\label{trajetory_control_vs_inpainting}
  \end{figure}

One of the features of the design of our hierarchical controller is that it can preserve the diversity of A-MDM, which allows the hierarchical model to produce a variety of different motions for performing the same task, as illustrated in Figure \ref{rl_target}. This is incontrast to the hierarchical models proposed by \citet{ling2020character}, which selects deterministic latents for steering the base model and therefore results in deterministic motions that sacrifice much of the diversty of the original base model. This experiment shows that our method can produce diverse motions while being controlled by a hierarchical controller for new tasks.

\subsection{Hierarchical Control vs Inpainting}
Hierarchical Control and Inpainting are able to perform identical tasks.  As shown in Figure \ref{trajetory_control_vs_inpainting}, with inpainting, the generated motion is restricted to exactly follow the user-specified trajectory. Therefore, if the target trajectory is unnatural, this can lead to unnatural full-body motions.  With hierarchical control, the generated motion has more flexibility to deviate from the specific target trajectory as needed to produce more natural behaviors at the cost of not following the target trajectory exactly. More examples are available in the supplementary video.

\subsection{A-MDM vs Space-time Diffusion Models}\label{vs_mdm}

Recent systems that apply diffusion models to motion synthesis have primarily focused on training space-time models for the conditional synthesis of motion sequences \cite{tevet2022human, karunratanakul2023gmd, zhang2022motiondiffuse}. This section highlights the key differences between A-MDM and space-time models. We start by comparing the performance of selected auto-regressive models and diffusion-based space-time models. In this experiment, we train our model with features provided by the data preprocessing pipeline of HumanML3D\citep{Guo_2022_CVPR}. We generally follow the evaluation pipeline of the main experiment of MDM\citep{tevet2022human} on HumanML3D. We measure Frechet Inception Distance (FID), diversity, and additionally foot skating ratio as in Guided Motion Diffusion (GMD) ~\citep{karunratanakul2023gmd}.

FID measures the difference between the distribution of the generated motion and that of the ground truth in learned latent space from a pre-trained motion model. Diversity computes the
average distance of generated motions in latent space.

During the evaluation, motions are generated with a fixed length of 196 frames for both auto-regressive and space-time models. Since auto-regressive models require an initial frame, we randomly sample this frame from the motion dataset. For space-time models, we sample motions unconditionally. The key difference between our evaluation and those conducted in the research of space-time models is that we emphasize the quality of unconditional generation, as opposed to text-conditioned generation. As demonstrated in Table\ref{spacetime}, A-MDM demonstrates the best performance among auto-regressive models. However, space-time models still demonstrate stronger performance across the metrics.
This difference may be due to A-MDM using a smaller number of diffusion steps than space-time models, and simpler network structures (MLP) for real-time synthesis). These factors result in significantly faster inference when evaluated under the standard DDPM sampling procedure without additional speedup techniques.

\begin{table}[h!]
\caption{Comparison between generative auto-regressive models and space-time models on HumanML3D~\citep{Guo_2022_CVPR}}\label{spacetime}
  \begin{tabular}{lccc}
    \toprule
     Model & FID$\downarrow$ & Diversity$\rightarrow$ &Foot Skat. Ratio$\downarrow$ \\
    \midrule
      Real & 0.002$\pm$0.00 & 9.5002 $\pm$ 0.002 & - \\
      \midrule
       MDM & 0.9157 $\pm$ 0.0533 &  9.0123 $\pm$ 0.0602 &  0.0930 $\pm$ 0.0021  \\
      GMD & \textbf{0.5727} $\pm$ 0.0681& \textbf{9.1714} $\pm$ 0.0789 &  \textbf{0.0657}$\pm$0.0016 \\
      \midrule
      MVAE & 11.2393$\pm$ 0.1607 & 6.1503 $\pm$ 0.0601 & 0.4153 $\pm$ 0.0025\\
      HuMoR & 8.2444$\pm$0.2437 & 7.7396 $\pm$ 0.0643 &  0.1210 $\pm$ 0.0011 \\
       \textbf{Ours} & 1.7435 $\pm$ 0.0813 & 7.8998 $\pm$ 0.0638 & 0.1010 $\pm$  0.0012 \\
    \bottomrule
  \end{tabular}
\end{table}

A-MDM and other auto-regressive models are trained to perform next-frame prediction. It can therefore generate long-horizon motions even when trained with a dataset that predominantly consists of short clips. In contrast, space-time models are generally trained with a maximum motion length, therefore using these models to generate longer motions is prone to drifting out of distribution from the training data. Please refer to the supplementary video for a comparison of this property.

\subsection{A-MDM vs Non-Generative Models}\label{vs_nsm}
To analyze A-MDM's capability to model diverse and highly dynamic motions, we compare our model with a non-generative auto-regressive model, Neural State Machine (NSM)\citep{starke2019neural}. We first evaluate the architecture and the general design between A-MDM and NSM by comparing the performance in an unconditional synthesis setting.
For the sake of a fair comparison, we directly use the same network as NSM and use phase information as the input for the gate network, which predicts the blending weights to the MoE framework. Additionally, we predict the phase value for the next time step in this model, as in NSM. We then train it on the same LaFAN1 dataset (without files concerning obstacles) as A-MDM. We compare the performance between A-MDM with NSM in Table \ref{nsm}. We found that our A-MDM effectively generates diverse motions with fewer artifacts than NSM. Results are available in the supplementary video. The difference in motion quality is particularly evident during a transition from a walk to a quick turn.

\begin{table}[h!]
\caption{Comparison between NSM and A-MDM. (unit: cm).}\label{nsm}
  \begin{tabular}{lccc}
    \toprule
     Model & FS$\downarrow$ & Pen.Freq$\downarrow$ &Pen.Dist$\downarrow$ \\
    \midrule
    GT & 1.58 & 0.79 $\%$ & 0.05\\
    \hline
       NSM &2.25 &  0.87 $\%$ &  0.07\\
       \textbf{Ours} & \textbf{1.99} & \textbf{0.82} $\%$ &  \textbf{0.04} \\
    \bottomrule
  \end{tabular}
\end{table}

\section{Discussion and Future Work}
In this work, we presented an auto-regressive diffusion model for kinematic motion synthesis. Unlike recent diffusion models for motion synthesis, which use space-models to generate motion sequences, we demonstrate that an auto-regressive diffusion model can synthesize high-quality and diverse motions, and can be effectively trained on large motion datasets.  To achieve real-time inference, we propose a lightweight architecture and use significantly fewer steps than conventional diffusion models. Once trained, A-MDM can be combined with a variety of different control methods to generate motions for new downstream tasks.

While our designs decisions improve overall stability and motion quality of A-MDM, training auto-regressive diffusion models for generate long-horizon motion is still challenging. A-MDM can occasionally exhibit unstable behaviours and failures under extreme circumstances. 
For instance, A-MDM is prone to generating motions that exhibit foot sliding and jittering when the user-specified controls are unnatural. 
We are interested in exploring techniques that can further mitigate these artifacts from A-MDM and improve the robustness of the model across different applications.
Our work has focused primarily on single-frame autoregressive models, but for some applications it may be beneficial to use multi-frame models, which predicts a sliding window of frames at a time. This may improve temporal coherence of the generated motions and allow the model tackle more complex tasks. Recent works have also proposed a variety of acceleration techniques for motion diffusion models \cite{zhou2023emdm,zhang2023tedi,song2023consistency}. Integrating these methods into A-MDM may further improve the runtime performance of these models for real-time applications.

% Bibliography
%%% -*-BibTeX-*-
%%% Do NOT edit. File created by BibTeX with style
%%% ACM-Reference-Format-Journals [18-Jan-2012].

\bibliographystyle{ACM-Reference-Format}
\bibliography{reference}
\end{document}